\documentclass[10pt,twocolumn,letterpaper]{article}

\usepackage{silence}
\usepackage{iccv}
\usepackage{times}
\usepackage{epsfig}
\usepackage{graphicx}
\usepackage{amsmath}
\usepackage{amssymb}

\usepackage{booktabs}
\usepackage{verbatim}
\usepackage{wrapfig,lipsum,booktabs}
\usepackage[dvipsnames]{xcolor}
\usepackage{mathtools} 
\usepackage{multirow}
\usepackage{subfigure}
\usepackage{graphicx} 
\usepackage{color}

\usepackage{arydshln}
\usepackage{minitoc}
\usepackage{algorithm2e}

\usepackage{comment}
\usepackage{siunitx}
\usepackage{float}
\usepackage{colortbl}
\usepackage{tabularx}
\usepackage{subfigure}

\makeatletter
\def\hlinewd#1{%
\noalign{\ifnum0=`}\fi\hrule \@height #1 \futurelet
\reserved@a\@xhline}

\usepackage{pifont}

\usepackage[pagebackref=true,breaklinks=true,colorlinks,bookmarks=false]{hyperref}
\iccvfinalcopy

\usepackage[capitalize]{cleveref}
\crefname{section}{Sec.}{Secs.}
\Crefname{section}{Section}{Sections}
\Crefname{table}{Table}{Tables}
\crefname{table}{Tab.}{Tabs.}
\newcommand{\figref}[1]{Fig.~\ref{#1}}
\newcommand{\tabref}[1]{Tab.~\ref{#1}}

\newcommand{\secref}[1]{Sec.~\ref{#1}}

\newcommand{\rom}[1]{\uppercase\expandafter{\romannumeral #1\relax}}
\newcommand{\paragrapht}[1]{\noindent\textbf{#1}}  
\newcommand{\ours}{Pano-I2I\xspace}

\newcommand{\contentx}{\mathbf{c}^\mathrm{x}}
\newcommand{\contentxaug}{\mathbf{c}^\mathrm{x}_\mathrm{aug}}
\newcommand{\contenty}{\mathbf{c}^\mathrm{y}}
\newcommand{\contentyhat}{\mathbf{c}^\mathrm{\hat{y}}}

\newcommand{\styley}{\mathbf{s}^\mathrm{y}}

\definecolor{turquoise}{cmyk}{0.65,0,0.1,0.3}
\definecolor{purple}{rgb}{0.65,0,0.65}
\definecolor{dark_green}{rgb}{0, 0.5, 0}
\definecolor{orange}{rgb}{0.8, 0.6, 0.2}
\definecolor{red}{rgb}{0.8, 0.2, 0.2}
\definecolor{darkred}{rgb}{0.6, 0.1, 0.05}
\definecolor{blueish}{rgb}{0.0, 0.3, .6}
\definecolor{blue}{rgb}{0, 0.3, 1}
\definecolor{light_gray}{rgb}{0.7, 0.7, .7}
\definecolor{pink}{rgb}{1, 0, 1}
\definecolor{greyblue}{rgb}{0.25, 0.25, 1}
\definecolor{light_gray}{gray}{0.95}
\definecolor{light-green}{rgb}{0.82, 0.94, 0.75}
\newcommand{\minus}{\text{-}}
\usepackage{array}
\usepackage{arydshln}

\begin{document}

\title{Panoramic Image-to-Image Translation}

\author{
Soohyun Kim\textsuperscript{1} \qquad Junho Kim\textsuperscript{2} \qquad Taekyung Kim\textsuperscript{2}  \qquad Hwan Heo\textsuperscript{1} \vspace{2.5pt}\\ \qquad Seungryong Kim\textsuperscript{1}\thanks{Corresponding authors} \qquad Jiyoung Lee\textsuperscript{2}\footnotemark[1] \qquad Jin-Hwa Kim\textsuperscript{2,3}\footnotemark[1] \ \vspace{2.5pt}\\\\
${}^{1}$Korea University \qquad ${}^{2}$NAVER AI Lab \qquad ${}^{3}$Seoul National University\\
}

\maketitle
\begin{abstract}
In this paper, we tackle the challenging task of Panoramic Image-to-Image translation (\ours) for the first time. This task is difficult due to the geometric distortion of panoramic images and the lack of a panoramic image dataset with diverse conditions, like weather or time.
To address these challenges, we propose a panoramic distortion-aware I2I model that preserves the structure of the panoramic images while consistently translating their global style referenced from a pinhole image.  
To mitigate the distortion issue in naive \ang{360} panorama translation, we adopt spherical positional embedding to our transformer encoders, introduce a distortion-free discriminator, and apply sphere-based rotation for augmentation and its ensemble. We also design a content encoder and a style encoder to be deformation-aware to deal with a large domain gap between panoramas and pinhole images, enabling us to work on diverse conditions of pinhole images. 
In addition, considering the large discrepancy between panoramas and pinhole images, our framework decouples the learning procedure of the panoramic reconstruction stage from the translation stage. 
We show distinct improvements over existing I2I models in translating the StreetLearn dataset in the daytime into diverse conditions. The code will be publicly available online for our community.

\end{abstract}

\section{Introduction}
\label{sec:intro}

\begin{figure}[t]
    \centering
  \includegraphics[width=1.0 \linewidth]{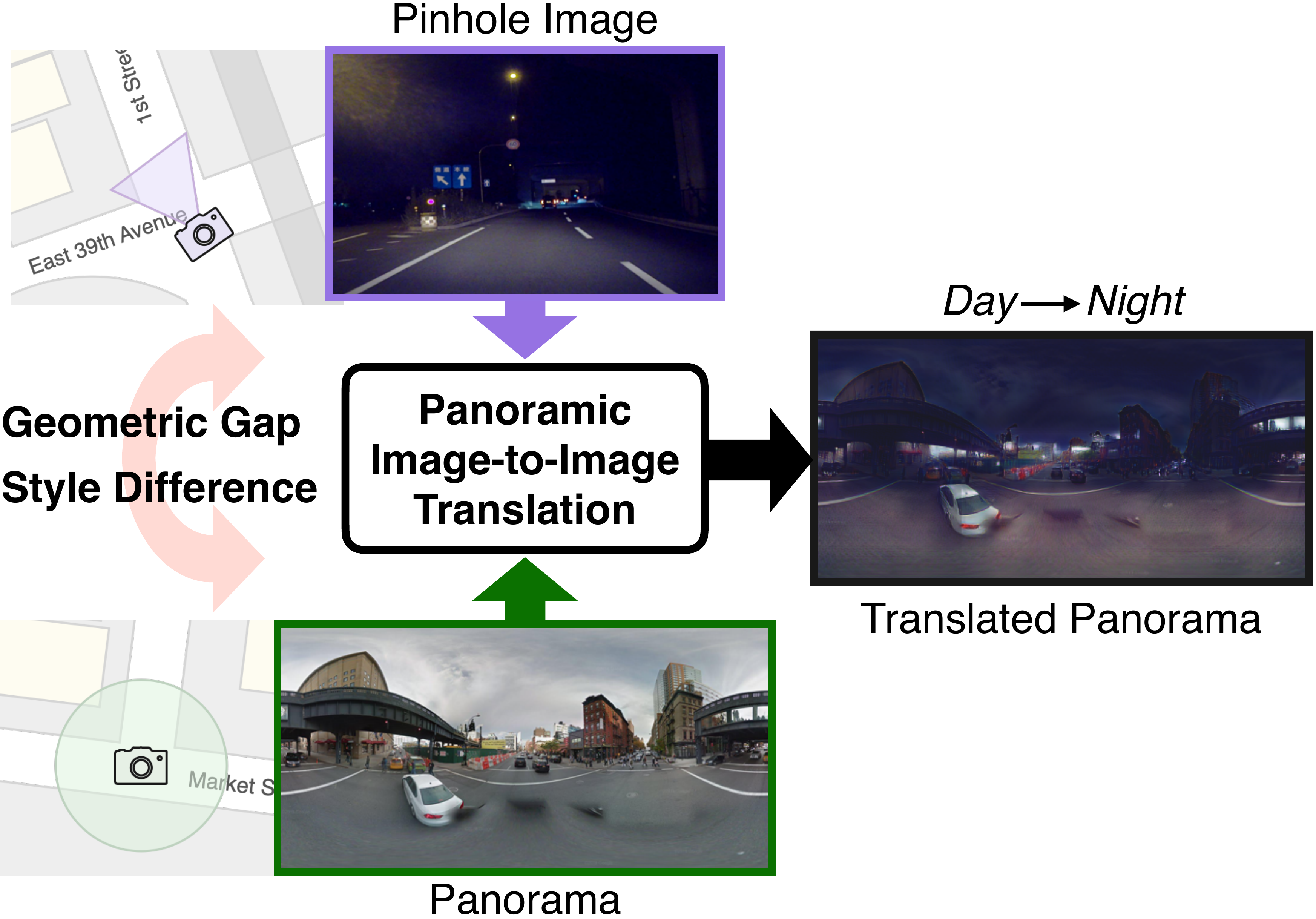}\hfill\\
  \caption{\textbf{Illustration of our problem formulation.}
Our \ours is trained on panoramas in the daytime as the source domain and pinhole images with diverse conditions as the target domain, where the two domains have significant geometric and style gaps.
}    \label{fig:formulation}
\end{figure}

Image-to-image translation (I2I) aims to modify an input image aligning with the style of the target domain, preserving the original content from the source domain. This paradigm enables numerous applications, such as colorization, style transfer, domain adaptation, data augmentation, \etc~\cite{huang2018auggan, mariani2018bagan,murez2018image,huang2017arbitrary,isola2017image}. However, existing I2I has been used to synthesize pinhole images with narrow field-of-view (FoV), which limits the scope of applications considering diverse image-capturing devices.

Panoramic \ang{360} cameras have recently grown in popularity, which enables many applications, \eg, AR/VR, autonomous driving, and city map modeling~\cite{micusik2009piecewise,anderson2018vision,caruso2015large,yogamani2019woodscape}.
Unlike pinhole images of narrow FoV, \textbf{panoramic images} (briefly, \textbf{panoramas}) capture the entire surroundings, providing richer information with \ang{360}$\times$\ang{180} FoV. 
Translating panoramas into other styles can enable novel applications, such as immersive view generations or enriching user experiences with robust car-surrounding recognition~\cite{de2018eliminating,yang2020ds,yang2021capturing,ma2021densepass}. 

However, naively applying conventional I2I methods for pinhole images~\cite{park2020contrastive,shen2019towards,choi2018stargan,choi2020stargan,zheng2021spatially,jeong2021memory,kim2022instaformer} to panoramas can significantly distort the geometric properties of panoramas as shown in \figref{fig:formulation}.
One may project the panoramic image into pinhole images to apply the conventional methods. However, it costs a considerable amount of computation since sparse projections cannot cover the whole scene due to the narrow FoV of pinhole images. In addition, the discontinuity problem at edges (left-right boundaries in panorama) requires panorama-specific modeling, as in the other tasks, \eg, panorama depth estimation, panoramic segmentation, and panorama synthesis~\cite{shen2022panoformer,zhang2022bending,chen2022text2light}.

Another challenge of panoramic image-to-image translation is the absence of sufficient panorama datasets.
Compared to the pinhole images, panoramic images are captured by a specially-designed camera (\ang{360} camera) or post-processed using multi-view images obtained from the calibrated cameras.
Especially for I2I, panoramas obtained under diverse conditions such as sunny, rainy, and night are needed to define the target or style domain.
Notice that panoramic images for the Street View service are mainly taken during the day~\cite{mirowski2018learning}.
Instead of constructing a new panorama dataset that is costly to obtain, it would be highly desirable if we could leverage existing pinhole image datasets with various conditions as style guidance.

\begin{figure}[t]
    \centering
   \includegraphics[width=1.0\linewidth]{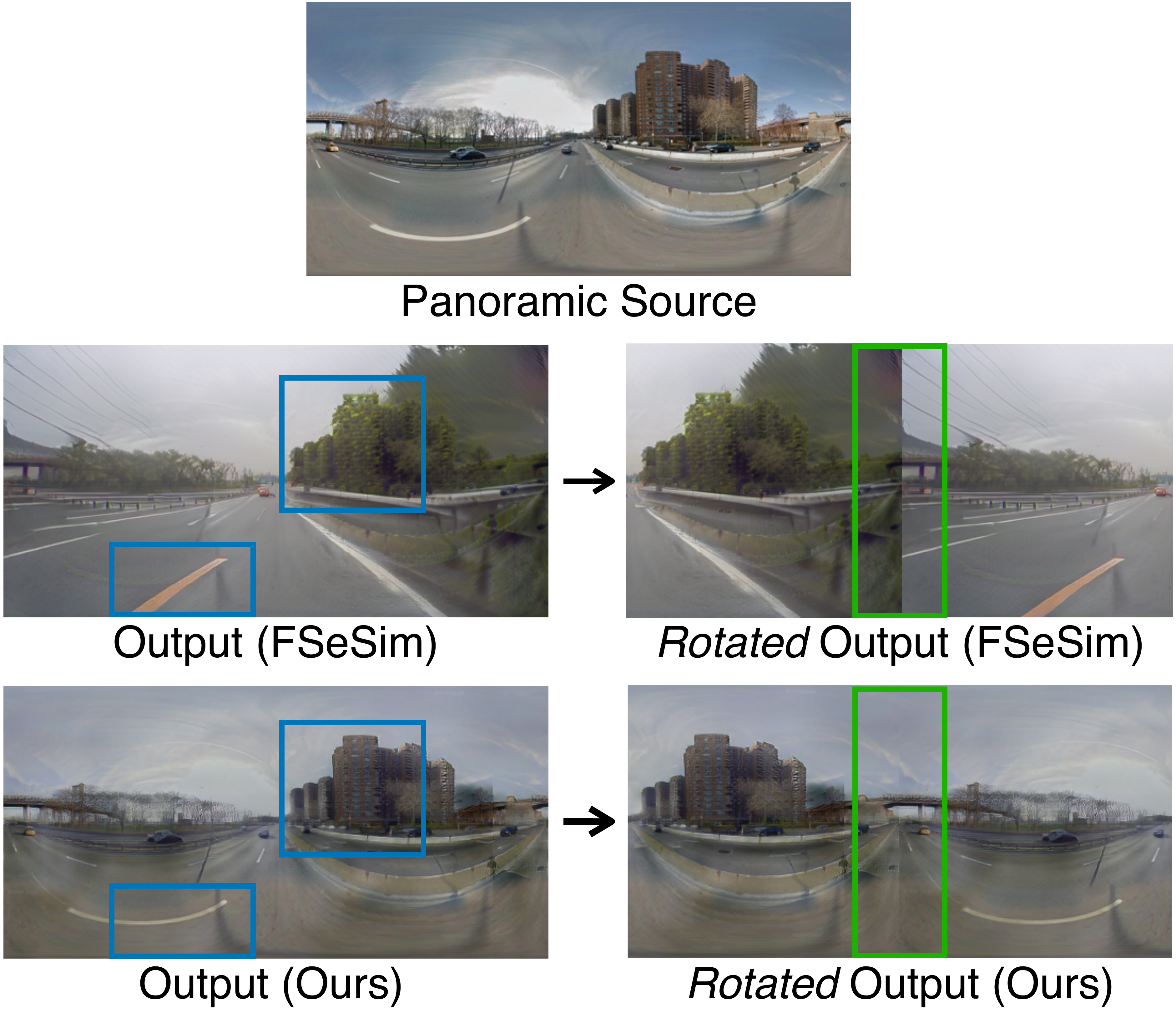}\hfill\\
\caption{\textbf{Typical failure cases of an existing method (FSeSim~\cite{zheng2021spatially}) for the panoramic image-to-image translation task.}
The generated image from FSeSim~\cite{zheng2021spatially} shows the collapsed result having a pinhole-like structure~(\textcolor{blue}{blue} box), as it overfits the pinhole target. Additionally, it has structural- and style-discontinuity in the edges~(\textcolor{green}{green} box). 
In contrast, our method generates high-quality panoramic images, achieving  a rotation-equivariant structure at the edges. We visualize rotated outputs ($\theta=\ang{180}$) to highlight the discontinuity.}
    \label{fig:failures}
\end{figure}

In summary, there are several challenges to translating panoramas into another condition: 1) the geometric deformation due to the wide FoV of panoramas, 2) distortion and discontinuity problems arising when existing methods are directly applied, and 3) the lack of panoramic image datasets with diverse conditions. We present typical failure cases
of existing approaches~\cite{zheng2021spatially} in \figref{fig:failures}. Based on the above analysis, we seek to expand the applicability of I2I to panoramic images by employing existing pinhole image datasets as style domain, dubbed the \textit{Panoramic Image-to-Image Translation}, shortly, \ours.

To address geometric deformation in panoramas, we adopt deformable convolutions~\cite{zhu2019deformable} to our encoders, with different offsets for panoramas and pinhole images to reflect the geometric differences between the source and target. To handle the large domain gap between the source and target domain, we propose a distortion-free discrimination that attenuates the effects of the geometric differences. In addition, we adopt panoramic rotation augmentation techniques to solve discontinuity problems at edges, considering that a $\ang{360}$ panorama should be continuous at boundaries.
Moreover, we propose a two-stage learning framework for stable training since learning with panorama and pinhole images simultaneously might increase the problem's complexity. Along with the Stage-\rom{1} that first performs fine-tuning on panoramas, the Stage-\rom{2} learns how to translate the panoramas attaining styles from pinhole images. 

We validate the proposed approach by evaluating the panorama dataset, StreetLearn~\cite{mirowski2018learning}, and day-and-night
 and weather conditions of the pinhole datasets~\cite{shen2019towards,sakaridis2019guided}. 
Our proposed method significantly outperforms all existing methods across various target conditions from the pinhole datasets. We also provide ablation studies to validate and analyze the components in \ours.

In summary, our main contributions are:
\begin{itemize}
    \item For the first time, to the best of our knowledge, we propose the panoramic I2I task and approach translating panoramas with pinhole images as a target domain.
    \item We present distortion-free discrimination to deal with a large geometric gap between the source and the target.
    \item Our spherical positional embedding and sphere-based rotation augmentation efficiently handle the geometric deformation and structural- and style-discontinuity at the edges of panoramas.
    \item \ours notably outperforms the previous methods in the quantitative evaluations for style relevance and structural similarity, providing qualitative analyses.

\end{itemize}

\section{Related work}
\label{sec:2}
\paragrapht{Image-to-image translation.}
\label{subsec:2.1}
Different from early works requiring paired dataset~\cite{isola2017image}, the seminal works~\cite{CycleGAN2017} enabled unpaired source/target training (\ie, learning without the ground-truth of the translated image).
Some works enable multimodal learning~\cite{huang2018multimodal,lee2018diverse,lee2020drit++}, multi-domain learning~\cite{choi2018stargan,choi2020stargan,wu2019relgan} for diverse translations from unpaired data, and instance-aware learning~\cite{shen2019towards,bhattacharjee2020dunit,jeong2021memory,kim2022instaformer} in complex scenes.
Nevertheless, existing I2I methods are restrictive to specific source-target pairs; they are limited to handling geometric variations (\eg, part deformation, viewpoint, and scale) between the source domain and the target domain. Our approach introduces a robust framework to an unpaired setting, even with geometric differences.
Also, the abovementioned methods may fail to obtain rotational equivalence for panorama I2I. 
On the other hand, several works have adopted the architecture of vision transformers~\cite{dosovitskiy2020image} to image generation~\cite{lee2021vitgan,jiang2021transgan,zhang2022styleswin}. Being capable of learning long-range interactions, the transformer is often employed for high-resolution image generation~\cite{esser2021taming,zhang2022styleswin}, or complex scene generation~\cite{wang2021sceneformer,kim2022instaformer}.
For instance, InstaFormer~\cite{kim2022instaformer} proposed to use transformer-based networks for I2I, capturing global consensus in complex street-view scenes.

\paragrapht{Panoramic image modeling.}
\label{subsec:2.2}
Panoramic images from \ang{360} cameras provide a thorough view of the scene with a wide FoV, beneficial in
understanding the scene holistically.
A common practice to address distortions in panoramas is to project an image into other formats of \ang{360} images (\textit{e.g.,} equirectangular, cubemap)~\cite{cheng2018cube,wang2018omnidirectional,yang2019dula}, and some works even combine both equirectangular and cubemap projections with improving performance~\cite{jiang2021unifuse,wang2020bifuse}.
However, they do not consider the properties of \ang{360} images, such as the connection between the edges of the images and the geometric distortion caused by the projection.
Several works leverage narrow FoV projected images~\cite{lee2018memory,yang2018object,de2018eliminating}, but they require many projected images (\eg, 81 images~\cite{lee2018memory}), which is an additional burden.
To deal with such discontinuity and distortion, recent works introduce modeling in spherical domain~\cite{esteves2018learning,cohen2018spherical}, projecting an image to local tangent patches with minimal geometric error.
It is proved that leveraging transformer architecture in \ang{360} image modeling reduces distortions caused by projection and rotation~\cite{cho2022spherical}. For this reason, recent approaches~\cite{ranftl2021vision,rey2022360monodepth} including PAVER~\cite{yun2022panoramic}, PanoFormer~\cite{shen2022panoformer}, and Text2Light~\cite{chen2022text2light} used the transformer achieving global structural consistency.

\label{subsec:2.3}

\begin{figure*}[t]
\begin{center}
{\includegraphics[width=1.0\linewidth]{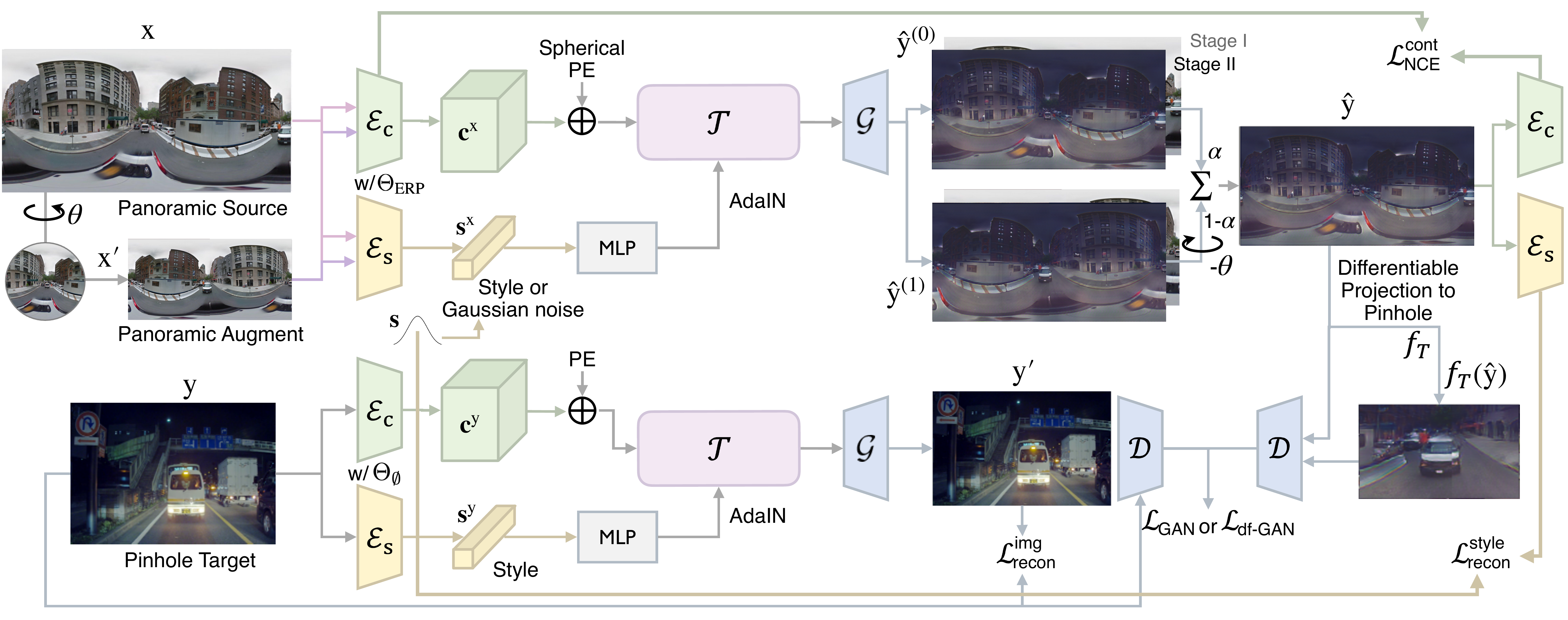}}\hfill\\ 
  \caption{\textbf{Overall network configuration of \ours}, consisting of content and style encoders, transformer, generator, and discriminator. Given panoramas as the source domain, we disentangle the content and translate its style aligned to the target domain. For training, pinhole datasets are used as targets referring to styles. Panoramic augmentation and ensemble are also introduced to preserve the spherical structure of panoramas. In our framework, Stage \rom{1} only learns to reconstruct the panorama source, and panorama translation is learned in Stage \rom{2}.}
\end{center}
  \label{fig:architecture}
\end{figure*}

\section{Methodology}

\subsection{Problem definition}\label{sec:3.1}
In our setting, we use the panoramic domain as a source that forms content structures and the pinhole domain as a target style. More formally, given a panorama of the source domain $\mathcal{X}$, \ours aims to learn a mapping function that translates its style into target pinhole domain $\mathcal{Y}$ retaining the content and structure of the panorama. Unlike the general I2I methods~\cite{CycleGAN2017,park2020contrastive,choi2018stargan,choi2020stargan,zheng2021spatially} that have selected source and target domains both in a narrow FoV condition, our setting varies both in style and structure: the source domain as panoramas with wide FoV, captured in the daytime and the target domain as pinhole images in diverse conditions with narrow FoV.
In this setting, existing state-of-the-art I2I methods~\cite{CycleGAN2017,park2020contrastive,park2020swapping,jiang2020tsit,choi2020stargan,liu2021smoothing,zheng2021spatially,jeong2021memory,kim2022instaformer} designed for pinhole images may fail to preserve the panoramic structure of the content, since their existing feature disentanglement methods cannot separate style from the target content because there exist both \textit{geometric} and \textit{style} differences between the source and the target domains. 
We empirically observed that 1) the outputs of the existing I2I method result in pinhole-like images, as shown in~\figref{fig:failures}, and 2) pinhole image-based network design that does not consider the spherical structure of \ang{360} causes discontinuity at left-right boundaries and low visual fidelity.

\subsection{Architecture design}\label{sec:3.3}
\paragrapht{Overall architecture.}
On a high level, the proposed method consists of a shared content encoder $\mathcal{E}_{\mathrm{c}}$, a shared style encoder $\mathcal{E}_{\mathrm{s}}$ to estimate the disentangled representation, a transformer encoder $\mathcal{T}$ to mix the style and content through AdaIN~\cite{huang2017arbitrary} layers, and a unified generator $\mathcal{G}$ and discriminator $\mathcal{D}$ to generate the translated image.
Our transformer encoder block consists of a multi-head self-attention layer and a feed-forward MLP with GELU~\cite{vaswani2017attention}.

In specific, to translate an image $\mathbf{x}$ in source domain $\mathcal{X}$ to target domain $\mathcal{Y}$, we first extract a content feature map $\contentx\in{\mathbb{R}}^{ h \times w \times l_{c}}$ by the content encoder from $\mathbf{x}$, with height $h$, width $w$, and $l_{c}$ content channels, and receive a random latent code $\mathbf{s} \in \mathbb{R}^{1 \times 1 \times {l_{s}}} $ from Gaussian distribution $\mathcal{N}(0, \mathbf{I}) \in \mathbb{R}^{l_s}$, which is used to control the style of the output with the affine parameters for AdaIN~\cite{huang2017arbitrary} layers in transformer encoders. Finally, we get the output image by $\mathbf{\hat{y}} = \mathcal{G(\mathcal{T}(\contentx,\mathbf{s}))}$. 
In the following, we will explain our key ingredients -- panoramic modeling in the content encoder, style encoder and the transformer, distortion-free discrimination, and sphere-based rotation augmentation and its ensemble  -- in detail.

\paragrapht{Panoramic modeling in encoders.}
In \ours, latent spaces are shared to embed domain-invariant content or style features as done in~\cite{lee2018diverse}.
Namely, the content and style encoders, $\mathcal{E}_\mathrm{c}$ and $\mathcal{E}_\mathrm{s}$, respectively, take either panoramas or pinhole images as inputs.
However, the geometric distortion gap from the different FoVs prevents the encoders from understanding each corresponding structural information.
To overcome this, motivated from~\cite{shen2022panoformer,yun2022panoramic}, we use the specially-designed deformable convolution layer~\cite{zhu2019deformable} at the beginning of the content and style encoders by adjusting the offset reflecting the characteristics of the image types. 
Specifically, given a panorama, the deformable convolution layer can be applied directly to the panorama by deriving an equirectangular plane (ERP) offset~\cite{yun2022panoramic}, $\Theta_{\text{ERP}}$, that considers the panoramic geometry.
To this end, we project a tangential local patch $\mathcal{P}$ of a 3D-Cartesian domain into ERP to obtain the corresponding ERP offset, $\Theta_{\text{ERP}}$, as follows:
\begin{equation}
    \Theta_{\text{ERP}}(\theta,\phi) = f_{\text{SPH}\rightarrow\text{ERP}}(f_{\text{3D}\rightarrow\text{SPH}}(\frac{\mathcal{P}\times R(\theta,\phi)}{||\mathcal{P}\times R(\theta,\phi)||_2})),
\end{equation}
where $R(\theta, \phi)$ indicates the rotation matrix with the longitude $\theta \in [0,2\pi]$ and latitude $\phi \in [0,\pi]$, $f_{\text{SPH}\rightarrow\text{ERP}}$ indicates the conversion function from spherical domain to ERP domain, and $f_{\text{3D}\rightarrow\text{SPH}}$ indicates the conversion function from 3D-Cartesian domain to spherical domain, as in \cite{yun2022panoramic}. 
To be detailed, we first project the tangential local patch $\mathcal{P}$ to the corresponding spherical patch on a unit sphere $S^{2}$, aligning the patch center to $(\theta, \phi )\in S^{2}$. Notice that the number of $\mathcal{P}$ is $H \times W$ with the stride of 1 and proper paddings, while the center of $\mathcal{P}$ corresponds to the kernel center.
Then, we obtain the relative spherical coordinates from the center point and all discrete locations in $\mathcal{P}$. Finally, these positions are projected to the ERP domain, represented as offset points. 
We compute such 2D-offset points $\Theta_\text{ERP} \in \mathbb{R}^{2\times H \times W \times \text{ker}_h \times \text{ker}_w}$ for each kernel location, and fixed to use them throughout training and test phase.

Unlike basic convolution, which has a square receptive field having limited capability to deal with geometric distortions in panoramas, our deformable convolution with fixed offsets can encode panoramic structure. 
We carefully clarify the objective of using deformable convolution is different from PAVER~\cite{yun2022panoramic}, which exploits the pinhole-based pretrained model for panoramic modeling. In contrast, \ours aims to learn both pinhole images and panorama in the shared networks simultaneously.
For pinhole image encoding, $\Theta_{\text{ERP}}$ is replaced to zero-offset $\Theta_{\varnothing}$ in both content and style encoders, which are vanilla convolutions. 

\paragrapht{Panoramic modeling in the transformer.}
After extracting the content features from the source image, we first patchify the content features to be processed through transformer blocks, then add positional embedding (PE)~\cite{vaswani2017attention,rahaman2019spectral}.
We represent the center coordinates of the previous patchified grids as $(i_\mathrm{p}, j_\mathrm{p})$ corresponding to the $p$-th patch having the width $w$ and the height $h$. As two kinds of inputs $\{\mathbf{x}$, $\mathbf{y}\}$ ---panorama, pinhole image, for each--- have different structural properties, we adopt the sinusoidal PE in two ways: using 2D PE and spherical PE (SPE), respectively.

To start with, we define the absolute PE in transformer~\cite{vaswani2017attention} as $\gamma(\cdot)$, a sinusoidal mapping into $\mathbb{R}^{2K}$ as
\begin{equation}
    \gamma(a) =\{ 
    (\mathrm{sin}(2^{k-1}\pi \textit{a}),\mathrm{cos}(2^{k-1}\pi \textit{a})) 
    | k=1,...,K\}
\end{equation}
for an input scalar $a$. Based on this, we define the 2D PE for common pinhole images as follows:

\begin{equation}
    \text{PE} = \mathrm{concat}(\gamma(i_\mathrm{p}),\gamma(j_\mathrm{p})).
\end{equation}
Following the previous work~\cite{chen2022text2light}, we consider the \ang{360} spherical structure of panorama presenting a spherical positional embedding for the center position ($i_\mathrm{p}$,$j_\mathrm{p}$) of each grid defined as follows, further added to the patch embedded tokens to work as explicit guidance:
\begin{equation}
\begin{split}
    \text{SPE} & = \mathrm{concat}(\gamma(\theta),\gamma(\phi)), \\
    \mathrm{where}~\theta = (2i_\mathrm{p}/&h-1)\pi,\,\phi = (2j_\mathrm{p}/w-1)\pi/2.
    \end{split}
\end{equation}
Since SPE explicitly provides cyclic spatial guidance and the relative spatial relationship between the tokens, it helps to maintain rotational equivariance for the panorama by encouraging structural continuity at boundaries in an $\ang{360}$ input. On the other hand, the previous spherical modeling methods used the standard learnable PE~\cite{yun2022panoramic,zhang2022bending} or limited to employ SPE as a condition for implicit guidance~\cite{chen2022text2light}, which does not provide token-wise spatial information. 
This positional embedding is added to patch-embedded tokens and further processed into transformer encoders with AdaIN. 

\paragrapht{Distortion-free discrimination.}
Contrary to the domain setting in traditional I2I methods that have only differences in style, our source and target domains exhibit two distinct features: geometric structure and style.
As shown in the blue box in \figref{fig:failures}, directly applying the existing I2I method for panoramic I2I guided by pinhole images brings severe structural collapse blocking artifacts, severely affecting the synthesizing quality.
We speculate that this problem, which has not been explored before, breaks the discriminator and causes structural collapse.
Concretely, while the discriminator in I2I typically learns to distinguish real data $\mathbf{y}$ and fake data $\mathbf{\hat{y}}$, mainly focusing on style difference, in our task, there is an additional large deformation gap and FoV difference between $\mathbf{y}$ and $\mathbf{\hat{y}}$ that confuses what to discriminate.

To address this issue, we present a distortion-free discrimination technique.
The key idea is to transform a randomly selected region of a panorama into a pinhole image while maintaining the degree of FoV.  
Specifically, we adopt a panorama-to-pinhole image conversion $f_T$ by a rectilinear projection~\cite{lee2018memory}. To obtain a narrow FoV (pinhole-like) image with $f_T$, we first select a viewpoint in the form of longitude and latitude coordinates ($\theta$, $\phi$) in the spherical coordinate system, where $\theta$ and $\phi$ are randomly selected from [0,2$\pi$] and [0,$\pi$], respectively, and extract a narrow FoV region from the \ang{360} panorama image by a differentiable projection function $f_T$. To further improve the discriminative ability of our model, we adopt a weighted sum of the original discrimination and the proposed discrimination to encourage the model to learn more robust features by considering both the original full-panoramas and pinhole-like converted panoramas.

\paragrapht{Sphere-based rotation augmentation and ensemble.}
We introduce a panoramic rotation-based augmentation since a different panorama view consistently preserves the content structure without the discontinuity problem at left-right boundaries. 
Given a panorama $\mathbf{x}$, a rotated image $\mathbf{x}'$ is generated by horizontally rotating $\theta$ angle. We efficiently implement this rotation by rolling the images in the ERP space without the burden of ERP$\rightarrow$SPH$\rightarrow$ERP projections since both are \textit{effectively} the same operation for the ERP domain.
The rotation angle is randomly sampled in [0, $2\pi$], where the step size is $2\pi$/10.
Such rotation is also reflected in SPE by adding the rotation angle $\theta$ to help the model learn the horizontal cyclicity of panoramas.
Later, the translated images $\mathbf{\hat{y}}^{(0)}$ and $\mathbf{\hat{y}}^{(1)}$ from the generator with $\mathbf{x}$ and $\mathbf{x}'$, respectively, are blended together, after rotating back with $-\theta$ for $\mathbf{\hat{y}}^{(1)}$ of course, to generate the final ensemble output $\mathbf{\hat{y}}$:
\begin{equation}
    \mathbf{\hat{y}}=\frac{\mathbf{\hat{y}}^{(0)} +\mathbf{\hat{y}}^{(1)'}}{2},
\end{equation}
where $\mathbf{\hat{y}}^{(1)'}$ is indicates $-\theta$ rotated version of $\mathbf{\hat{y}}^{(1)}$.
Thus, the result $\mathbf{\hat{y}}$ has more smooth boundary than the results predicted alone, mitigating discontinuous edge effects.

\subsection{Loss functions}\label{sec:3.4}

\paragrapht{Adversarial loss} minimizes the distribution discrepancy between two different features~\cite{goodfellow2014generative,mirza2014conditional}. We adopt this to learn the translated image $\mathbf{\hat{y}} = \mathcal{G(\mathcal{T}(\mathcal{E}_{\mathrm{c}}(\mathbf{x}, \mathrm{\Theta}_{\text{ERP}}),\mathbf{s}))}$ and the image $\mathbf{x}$ from $\mathcal{X}$ to have indistinguishable distribution to preserve panoramic contents, defined as:
\begin{equation}
\begin{split}
     \mathcal{L}_{\mathrm{GAN}} = &\mathbb{E}_{\mathbf{x}\sim\mathcal{X}}[\mathrm{log}(1-\mathcal{D}(\hat{\mathbf{y}}))]+ \mathbb{E}_{\mathbf{y} \sim\mathcal{Y}}[\mathrm{log}\, \mathcal{D}(\mathbf{y})],
\end{split}
\end{equation}
with the R1 regularization~\cite{mescheder2018training} to enhance training stability.
To consider the panoramic distortion-free discrimination using the panorama-to-pinhole conversion $f_T$, we define additional adversarial loss as follows: 
\begin{equation}
\begin{split}
     \mathcal{L}_{\mathrm{df}\minus\mathrm{GAN}} = \mathbb{E}_{\mathbf{x}\sim\mathcal{X}}[\mathrm{log}(1-\mathcal{D}(f_T(\hat{\mathbf{y}})))] +\mathbb{E}_{\mathbf{y} \sim\mathcal{Y}}[\mathrm{log}\, \mathcal{D}(\mathbf{y})].
\end{split}
\end{equation}

\paragrapht{Content loss.}
To maintain the content between the source image $\mathbf{x}$ and translated image $\hat{\mathbf{y}}$, we exploit the spatially-correlative loss~\cite{zheng2021spatially} to define a content loss, with an augmented source $\mathbf{x}_\mathrm{aug}$. To get $\mathbf{x}_\mathrm{aug}$, we apply structure-preserving transformations to $\mathbf{x}$. This helps preserve the structure and learn the spatially-correlative map~\cite{zheng2021spatially} based on patchwise infoNCE loss~\cite{oord2018representation}, since it captures the domain-invariant structure representation.
Denoting that $\hat{\mathbf{v}}$ as spatially-correlative map of the query patch from $\contentyhat=\mathcal{E}_{\mathrm{c}}({\mathbf{\hat{y}}}, \Theta_{\text{ERP}})$, we pick the pseudo-positive patch sample $\mathbf{v}^{+}$ from $\contentx=\mathcal{E}_{\mathrm{c}}(\mathbf{x}, \Theta_{\text{ERP}})$ in the same position of the query patch $\hat{\mathbf{v}}$, and the negative patches $\mathbf{v}^{-}$ from the other positions of $\contentxaug$ and $\contentx$, except for the position of query patches $\hat{\mathbf{v}}$.
We first define a score function $\ell(\cdot)$ at the $l$-th convolution layer in $\mathcal{E}_\mathrm{c}$:
\begin{equation}
\begin{split}
    \ell(\hat{\mathbf{v}_l}, &\mathbf{v}^{+}_l, \mathbf{v}^{-}_l) =\\
    &\mathrm{-log}\left[
    \frac{\mathrm{exp}(\hat{\mathbf{v}_l}\cdot \mathbf{v}^{+}_l/\tau)}{
    \mathrm{exp}(\hat{\mathbf{v}_l}\cdot \mathbf{v}^{+}_l/\tau) + \sum_{\mathrm n=1}^{\mathrm N}\mathrm{exp}(\hat{\mathbf{v}_l}\cdot \mathbf{v}^{-}_{\mathrm n}/\tau)}\right],
\end{split}
\end{equation}
where $\tau$ is a temperature parameter. 
Then, the overall content loss function is defined as follows:
\begin{equation}
\begin{split}
    \mathcal{L}_\mathrm{NCE}^{\mathrm{cont}} = \mathbb{E}_{\mathbf{x}\sim\mathcal{X}}\sum_{l}\sum_{s}\ell(\hat{\mathbf{v}}_l(s), &\mathbf{v}^{+}_l(s), \mathbf{v}^{-}_l(S{\setminus}s)),
\end{split}
\end{equation}
where the index $s\in \{1,2,...,S_{l}\}$ and $S_{l}$ is a set of patches in each $l$-th layer, and $S{\setminus}s$ indicates the indices except $s$.

\paragrapht{Image reconstruction loss.} 
We additionally use the image reconstruction loss to enhance the disentanglement between content and style in a manner that our $\mathcal{G}$ can reconstruct an image for domain $\mathcal{Y}$. To be specific, $\mathbf{y}$ is fed into content encoder $\mathcal{E}_{\mathrm{c}}$ and style encoder $\mathcal{E}_{\mathrm{s}}$ to obtain a content feature map $\contenty=\mathcal{E}_\mathrm{c}(\mathbf{y}, \Theta_{\varnothing})$ and a style code $\styley=\mathcal{E}_\mathrm{
s}(\mathbf{y}, \Theta_{\varnothing})$. 
We then compare the reconstructed image ${\mathcal{G}(\mathcal{T}(\contenty,\styley))}$ with $\mathbf{y}$ as follows:
\begin{equation}
    \mathcal{L}_\mathrm{recon}^\mathrm{img} = \, \mathbb{E}_{\mathbf{y} \sim \mathcal{Y}}[\|{\mathcal{G}(\mathcal{T}(\contenty,\styley)) -\,\mathbf{y}}\|_{1}].
\end{equation}

\paragrapht{Style reconstruction loss.}
In order to better learn disentangled representation, we compute L1 loss between the style code from the translated image and input panorama,
\begin{equation}
\begin{split}
    \mathcal{L}^\mathrm{style}_{\mathrm{\mathrm{ref}\minus \mathrm{recon}}} = \mathbb{E}_{\mathbf{x}\sim\mathcal{X} }[\|
    {\mathcal{E}_\mathrm{s}(\hat{\mathbf{y}}, \Theta_{\text{ERP}})
    -\mathcal{E}_\mathrm{s}(\mathbf{x}, \Theta_{\text{ERP}})}
    \|_{1}].\vspace{-10pt}
\end{split}
\end{equation}
We also define the style reconstruction loss to reconstruct the style code $\mathbf{s}$, which is used for the generation of $\hat{\mathbf{y}}$.  Note that the style code $\mathbf{s}$ is randomly sampled from Gaussian distribution, not extracted from an image.
\begin{equation}
\begin{split}
    \mathcal{L}^\mathrm{style}_{\mathrm{\mathrm{rand}\minus \mathrm{recon}}} = \mathbb{E}_{\mathbf{x}\sim\mathcal{X} ,\mathbf{y}\sim\mathcal{Y}}[\|{\mathcal{E}_\mathrm{s}(\hat{\mathbf{y}}, \Theta_{\text{ERP}})
    -\,{\mathbf{s}}}\|_{1}].\vspace{-10pt}
\end{split}
\end{equation}

\subsection{Training strategy}\label{sec:3.5}

\paragrapht{Stage \rom{1}: Panorama reconstruction.}
To our knowledge, there is no publicly-available large-scale outdoor panorama data, especially captured in various weather or season conditions.
For this reason, we cannot use panoramas as a style reference.
In addition, in order to share the same embedding space in content and style, the network must be able to process pinhole images and panoramas simultaneously.

For the stable training of \ours, the training procedure is split into two stages  corresponding with different objectives.
In Stage \rom{1}, we pretrain the content and style encoders $\mathcal{E}_\mathrm{c,s}$, transformer $\mathcal{T}$, generator $\mathcal{G}$, and discriminator $\mathcal{D}$ using the panorama dataset only. 
Given a panorama, the parameters of our network are optimized to reconstruct the original with adversarial and content losses, and style reconstruction loss.
As the network learns to reconstruct the input self again, we use the style feature represented by a style encoder instead of the random style code. In addition, for $\mathcal{L}_{\mathrm{GAN}}$ in Stage \rom{1}, the original discriminator receives $\mathbf{x}$ instead of $\mathbf{y}$ as an input.
The total objective in Stage \rom{1} as follows:
\begin{equation}
\begin{split}
    \mathcal{L}_{\text{Stage\rom{1}}} = &\mathcal{L}_{\mathrm{GAN}}
+\lambda_{\mathrm{cont}}\mathcal{L}_\mathrm{NCE}^{\mathrm{cont}} 
+\lambda_\mathrm{style}\mathcal{L}_{\mathrm{ref}\minus \mathrm{recon}}^\mathrm{style},
\end{split}
\end{equation}
where $\lambda_{\{*\}}$ denotes balancing hyperparameters that control the importance of each loss.

\paragrapht{Stage \rom{2}: Panoramic I2I guided by pinhole image.}
In Stage \rom{2}, the whole network is fully trained with robust initialization by Stage \rom{1}.
Compared to Stage \rom{1}, panorama and pinhole datasets are all used in this stage.
Concretely, the main difference is that; (1) original discrimination is combined with our distortion-free discrimination as a weighted sum, (2) the style code is sampled from the Gaussian distribution to translate the panorama, (3) the panoramic rotation-based augmentation and its ensemble technique are leveraged to enhance the generation quality. 
Therefore, the total objective in Stage \rom{2} is defined as:
\begin{equation}
\begin{split}
    \mathcal{L}_{\text{Stage\text{2}}} = & \lambda_{\mathrm{df}\minus\mathrm{GAN}}\mathcal{L}_{\mathrm{df}\minus\mathrm{GAN}}
    +(1-\lambda_{\mathrm{df}\minus\mathrm{GAN}})\mathcal{L}_{\mathrm{GAN}} 
\\&+\lambda_{\mathrm{cont}}\mathcal{L}_\mathrm{NCE}^{\mathrm{cont}} 
+\lambda_\mathrm{style}\mathcal{L}_{\mathrm{rand}\minus \mathrm{recon}}^\mathrm{style}
+\lambda_\mathrm{recon}\mathcal{L}_\mathrm{recon}^\mathrm{img}.
\end{split}
\end{equation}
Notice that $\lambda_{\{*\}}$ is differently set to each stage, and please refer to Appendix \ref{supp:sec1}.

\begin{figure*}
  \centering
  \renewcommand{\thesubfigure}{}
    \subfigure[]
{\includegraphics[width=  0.3\linewidth]{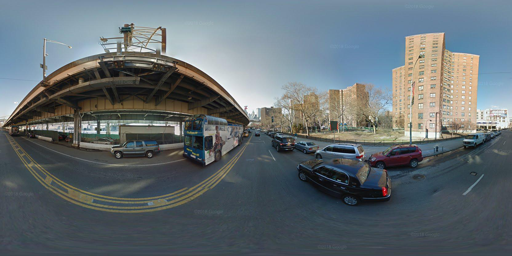}}  
    \subfigure[]
{\includegraphics[width=  0.3\linewidth]{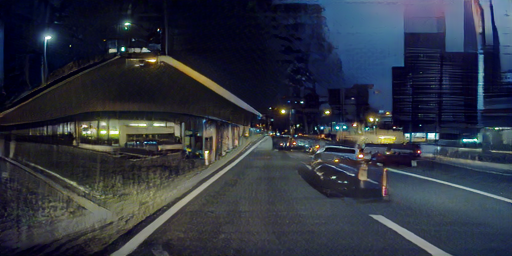}}  
    \subfigure[]
{\includegraphics[width=  0.3\linewidth]{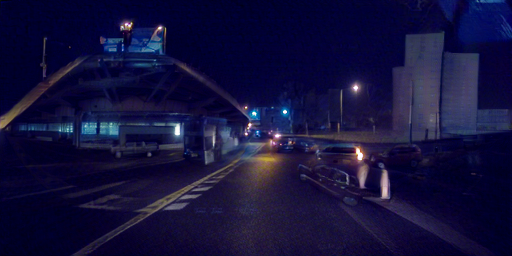}}  \\\vspace{-20.5pt}

\subfigure[Inputs]
{\includegraphics[width=  0.3\linewidth]{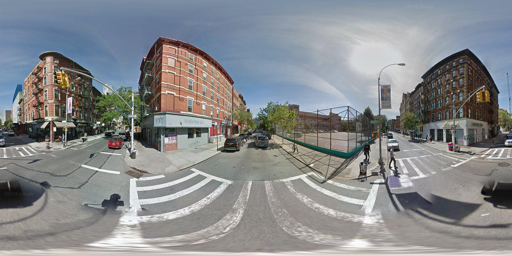}}  
\subfigure[CUT~\cite{park2020contrastive}]
{\includegraphics[width=  0.3\linewidth]{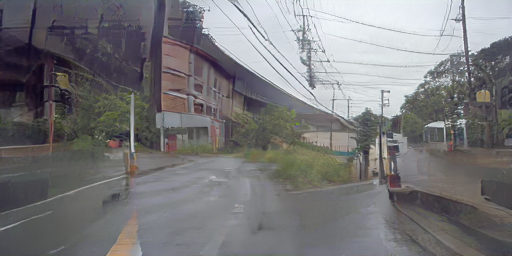}}  
\subfigure[FSeSim~\cite{zheng2021spatially}]
{\includegraphics[width=  0.3\linewidth]{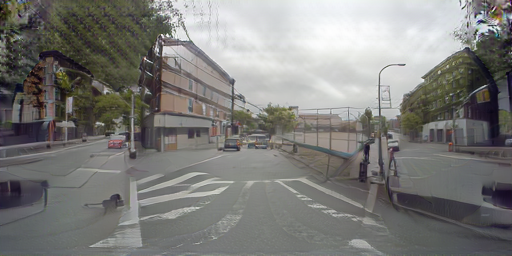}}  \\\vspace{-5pt}

\subfigure[]
{\includegraphics[width=  0.3\linewidth]{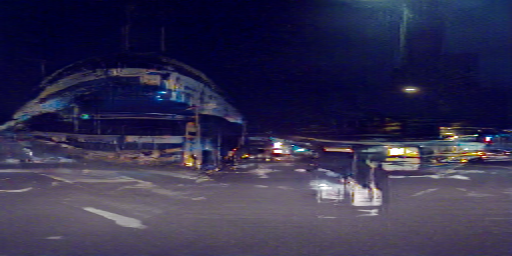}}  
\subfigure[]
{\includegraphics[width=  0.3\linewidth]{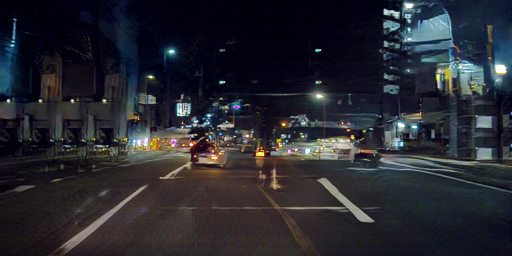}}  
\subfigure[]
{\includegraphics[width=  0.3\linewidth]{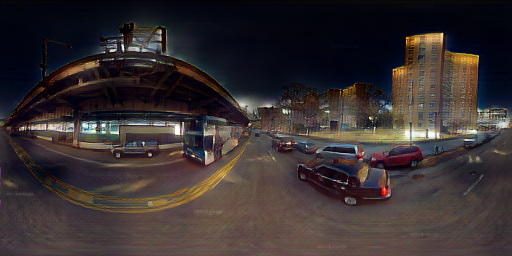}}  \\\vspace{-20.5pt}

\subfigure[MGUIT~\cite{jeong2021memory}]
{\includegraphics[width=  0.3\linewidth]{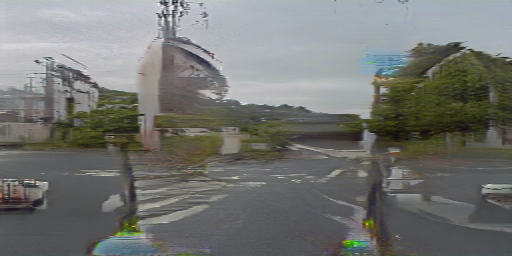}}  
\subfigure[InstaFormer~\cite{kim2022instaformer}]
{\includegraphics[width=  0.3\linewidth]{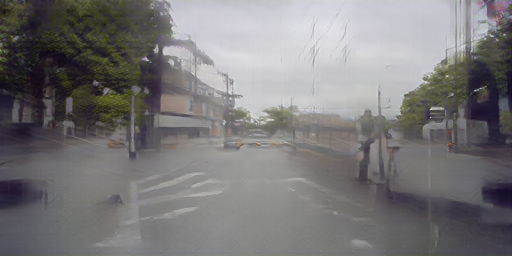}}  
\subfigure[Pano-I2I (ours)]
{\includegraphics[width=  0.3\linewidth]{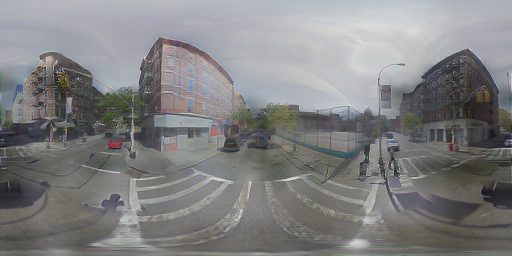}}  \\
\caption{\textbf{Qualitative comparison} on StreetLearn dataset (day) to INIT dataset (night, rainy): (top to bottom) day$\rightarrow$night, and day$\rightarrow$rainy results.
Among the methods, \ours (ours) preserves object details well and shows realistic results.}
  \label{fig:qualitative}
\end{figure*}

\section{Experiments}

\subsection{Experimental setup}
\paragrapht{Datasets.}
We conduct experiments on the panorama dataset, StreetLearn~\cite{mirowski2018learning}, as the source domain, and a standard street-view dataset for I2I, INIT~\cite{shen2019towards} and Dark Zurich~\cite{sakaridis2019guided}, as the target domain. 
StreetLearn provides \ang{360} outdoor 56k Manhattan panoramas taken from the Google Street View. 
Although INIT consists of four conditions (sunny, night, rainy, and cloudy), we use two conditions, night and rainy, since the condition of the StreetLearn is captured during the daytime, including sunny and cloudy. We use the \textit{Batch1} of the INIT dataset, a total of 62k images for the four conditions. Dark Zurich has three conditions (daytime, night, and twilight), a total of 8779 images, and we use night and twilight.

\paragrapht{Metrics.}
For quantitative comparison, we report the Fréchet Inception Distance (FID) metric~\cite{heusel2017gans} to evaluate style relevance, and the structural similarity (SSIM) index~\cite{wang2004image} metric to evaluate the panoramic content preserving.  
Considering that the structure of outputs tends to become pinhole-like in panoramic I2I tasks, we measure the FID metric after applying panorama-to-pinhole projection ($f_T$) for randomly chosen horizontal angle $\theta$ and fixed vertical angle $\phi$ as 0 with a fixed FoV of \ang{90}, for consistent viewpoint with the target images. 
Notice that the SSIM mediately shows the degree of content preservation because it measures the structural similarity between the original panorama and the translated panorama based on luminance, contrast and structure.

\paragrapht{Comparison methods.}
We compare our approach against the state-of-the-art I2I methods, including MGUIT \cite{jeong2021memory} and InstaFormer \cite{kim2022instaformer}, CUT \cite{park2020contrastive}, and FSeSim \cite{zheng2021spatially}. 
Since MGUIT and InstaFormer require bounding box annotations to train their models, we exploit pretrained YOLOv5~\cite{glenn_jocher_2021_4679653} model to generate pseudo bounding box annotations.

\subsection{Implementation details} 
We summarize the implementation details in the \ours.
We formulate the proposed method with vision transformers~\cite{dosovitskiy2020image} inspired by InstaFormer~\cite{kim2022instaformer}, but without instance-level approaches due to the absence of ground-truth bounding box annotations.
In training, we use the Adam optimizer~\cite{kingma2014adam} with $\beta_1$ = 0.5, $\beta_2$ = 0.999. The input of the network is resized into 256 $\times$ 512. We design our content and style encoders, a transformer encoder, the generator-and-discriminator for our GAN losses based on \cite{kim2022instaformer}, where all modules are learned from scratch. The initial learning rate is 1e-4, and the model is trained on 8 Tesla V100 with batch size 8 for Stage \rom{1} and 4 for Stage \rom{2}.

\begin{table}[t!]
\centering
\small
\begin{tabular}{l c c c c}
\toprule
\multirow{2}[1]{*}{Methods}
& \multicolumn{2}{c}{ Day$\rightarrow$Night} &  \multicolumn{2}{c}{Day$\rightarrow$Rainy} \\

\cmidrule(lr){2-3} \cmidrule(lr){4-5} 
& FID$\downarrow$ & SSIM$\uparrow$ & FID$\downarrow$ & SSIM$\uparrow$ \\ 
\midrule
CUT~\cite{park2020contrastive}     &  131.3 & 0.232 & 119.8 & 0.439 \\
FSeSim~\cite{zheng2021spatially} &  106.0 & 0.309 & 110.3 & 0.541 \\
MGUIT~\cite{jeong2021memory}       &  129.9 & 0.156 & 141.5 & 0.268\\
InstaFormer~\cite{kim2022instaformer} & 151.1 & 0.201 & 136.2 &0.495\\
\midrule
\ours (ours) &  
\textbf{94.3} & \textbf{0.417} & \textbf{86.6} & \textbf{0.708} \\
\bottomrule
\end{tabular}
\vskip 0.1in
\caption{\textbf{Quantitative evaluation} on the translated panoramas from the StreetLearn dataset to the INIT dataset.
}
\label{tab:main}
\end{table}

\subsection{Experimental results}
\paragrapht{Qualitative evaluation.}
In \figref{fig:qualitative}, we compare our method with other I2I methods.
We observe all the other methods ~\cite{park2020contrastive,jiang2020tsit,zheng2021spatially,jeong2021memory,kim2022instaformer} fail to synthesize reasonable panoramic results and show obvious inconsistent output regarding either structure or style in an image.
Moreover, previous methods recognize structural discrepancies between source and target domains as style differences, indicating failed translation results that change like pinhole images.
Surprisingly, in the case of `day$\rightarrow$night', all existing methods fail to preserve the objectness as a car or building. We conjecture that they can hardly deal with the large domain gap in `day$\rightarrow$night,' thus naively learning to follow the target distribution without considering the context from the source.
By comparison, our method shows the overall best performance in visual quality, preserving panoramic content, and structural- and style-consistency. Especially, we can observe the ability of our discrimination design to generate distortion-tolerate outputs. 
The qualitative results on Dark Zurich are provided in Appendix \ref{supp:sec5}.

\begin{table}[h]
\centering
\small
\begin{tabular}{l c c c c}
\toprule
\multirow{2}[1]{*}{Methods}
& \multicolumn{2}{c}{ Day$\rightarrow$Night} &  \multicolumn{2}{c}{ Day$\rightarrow$Twilight} \\

\cmidrule(lr){2-3} \cmidrule(lr){4-5} 
& FID$\downarrow$ & SSIM$\uparrow$ & FID$\downarrow$ & SSIM$\uparrow$ \\ 
\midrule
FSeSim~\cite{zheng2021spatially} & 133.8 & 0.305 & 138.8 & 0.420 \\
MGUIT~\cite{jeong2021memory} & 205.3 & 0.156 &229.9 & 0.124\\
\midrule
\ours (ours) &  \textbf{120.2} & \textbf{0.431} & \textbf{126.6} & \textbf{0.520} \\
\bottomrule
\end{tabular}
\vskip 0.1in
\caption{\textbf{Quantitative evaluation} on the translated panoramas from the StreetLearn dataset to the Dark Zurich dataset.
}
\label{tab:dz}
\end{table}

\begin{figure}[t]
  \begin{center}
\renewcommand{\thesubfigure}{}
    \subfigure[]
{\includegraphics[width=0.9\linewidth]{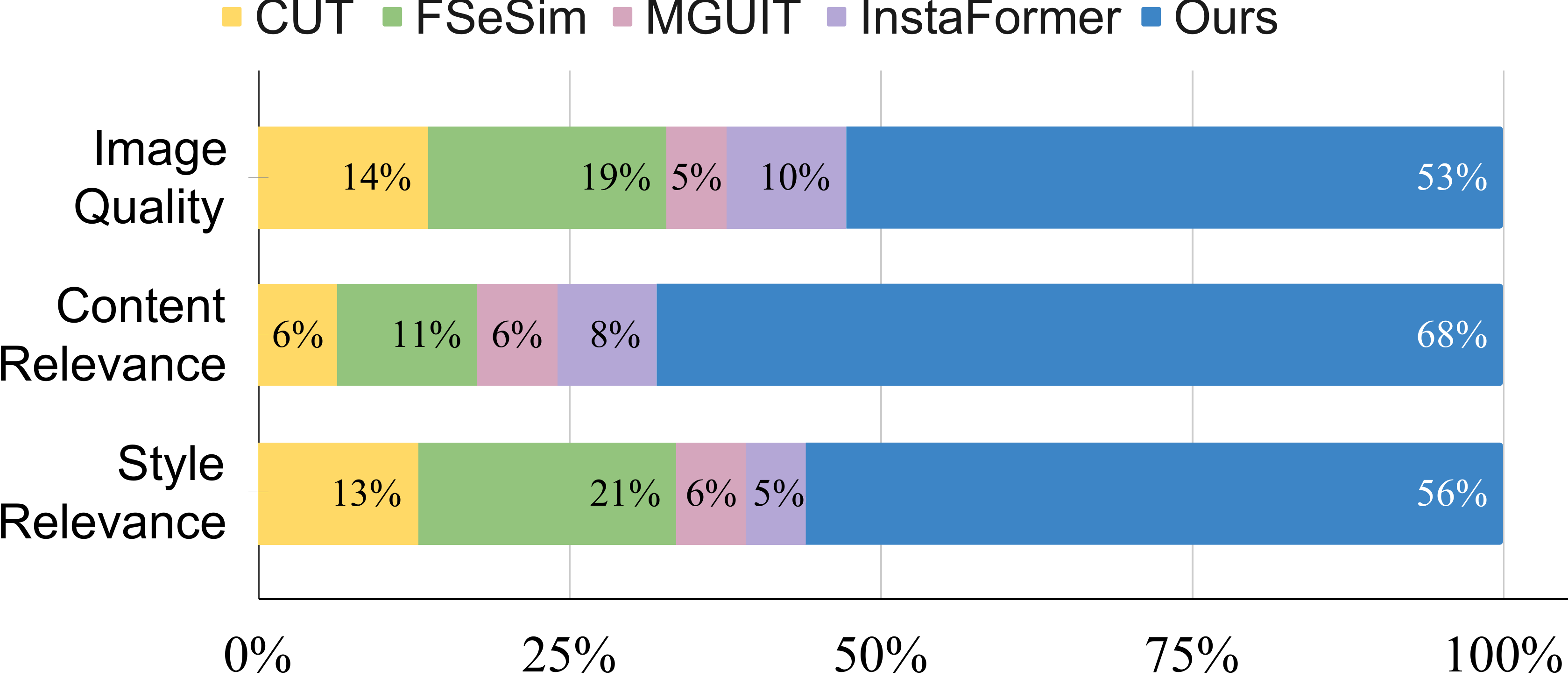}}\hfill\\\vspace{-10pt}
    \caption{\textbf{User study results.}
    }\label{fig:userstudy}
    \end{center}\vspace{-5pt}
\end{figure}

\paragrapht{Quantitative evaluation.}
\tabref{tab:main} and \tabref{tab:dz} show the quantitative comparison in terms of FID~\cite{heusel2017gans} and SSIM~\cite{wang2004image} index metrics. 
Our method consistently outperforms the competitive methods in all metrics, demonstrating that \ours successfully captures the style of the target domain while preserving the panoramic contents. Notably, our approach exhibits significant improvements in terms of SSIM.
In contrast, previous methods perform poorly in terms of SSIM compared to our results, which is also evident from the qualitative results presented in \figref{fig:qualitative}.

\paragrapht{User study.}
We also conduct a user study to compare the subjective quality. We randomly select 10 images for each task (sunny$\rightarrow$night, sunny$\rightarrow$rainy) on the INIT dataset, and let 60 users sort all the methods regarding ``overall image quality'', ``content preservation from the source'', and ``style relevance with the target, considering the context from the source''. As seen in \figref{fig:userstudy}, our method has a clear advantage on every task. We provide more details in Appendix \ref{supp:sec6}.

\subsection{Ablation study}
\begin{figure}[!t]
  \centering
\renewcommand{\thesubfigure}{}
    \subfigure[Input]
{\includegraphics[width=0.49\linewidth]{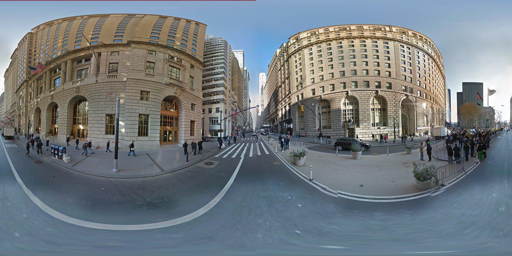}}
    \subfigure[(\rom{1}) Pano-I2I (ours)]
{\includegraphics[width=0.49\linewidth]{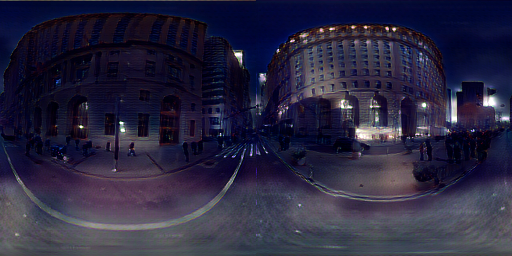}} \\ \vspace{-5pt}
    \subfigure[(\rom{2}) - Distortion-free $\mathcal{D}$]
{\includegraphics[width=0.49\linewidth]{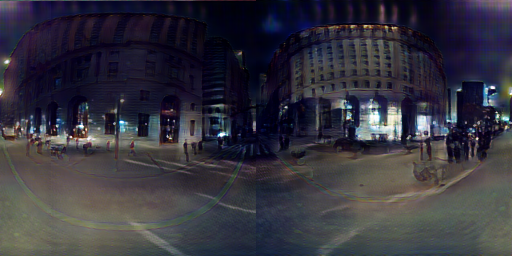}}
    \subfigure[(\rom{3}) - Ensemble]
{\includegraphics[width=0.49\linewidth]{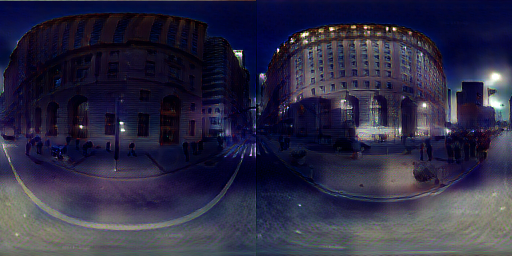}} \\\vspace{-5pt}
    \subfigure[(\rom{4}) - Two-stage learning]
{\includegraphics[width=0.49\linewidth]{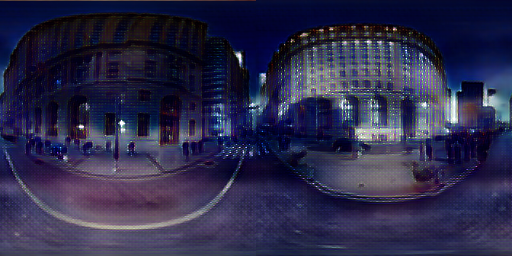}} 
\subfigure[(\rom{5}) - SPE, deform conv]
{\includegraphics[width=0.49\linewidth]{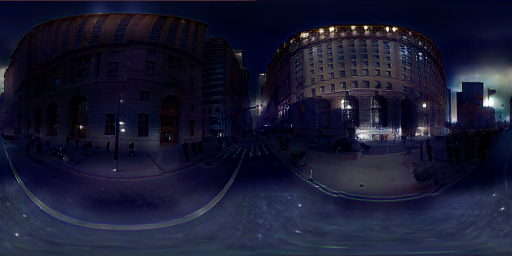}}
    \\
\caption{\textbf{Qualitative evaluation on ablation study.}}
  \label{fig:ablation}
\end{figure}

In \figref{fig:ablation} and \tabref{tab:abl}, we show qualitative and quantitative results for the ablation study on the day$\rightarrow$night task on the INIT dataset. 
In particular, we analyze the effectiveness of our 1) distortion-free discrimination, 2) ensemble technique, 3) two-stage learning scheme, and 4) spherical positional embedding (SPE) and deformable convolution.

As seen in \figref{fig:ablation}, our full model smoothens the boundary with high-quality generation, successfully preserving the panoramic structure. We also observe the ability of our discrimination design to generate distortion-tolerated outputs. The result without an ensemble fails to alleviate the discontinuity problem, as seen in the middle area of the image. The result without two-stage learning shows the limited capability to reconstruct the fine details of the contents from the input image. 
Since SPE and deformable convolution help the model learn the deformable structure of panoramas, the result without them fails to preserve the detailed structure. Note that the results are visualized after rotation ($\theta=\ang{180}$) to highlight the discontinuity.

In \tabref{tab:abl}, we measure the SSIM and FID scores to evaluate structural consistency and style relevance with respect to the choices of components. It demonstrates that our full model preserves input structure with our proposed components. 
We observe all the techniques and components contribute to improving the performance in terms of style relevance and content preservation, and the impact of distortion-free discrimination is substantially effective to handle geometric deformation.

\begin{table}[t]
\setlength{\tabcolsep}{8pt}
\small
\centering
\begin{tabular}{llccc}
\toprule
ID & Methods & FID$\downarrow$ & SSIM$\uparrow$ \\
\midrule
(\rom{1}) & {Pano-I2I (ours)} & \textbf{94.3} & \textbf{0.417} \\ \midrule
(\rom{2}) &\;(\rom{1}) {- Distortion-free $\mathcal{D}$} & 105.6 & 0.321 \\
(\rom{3}) &\;(\rom{1}) {- Ensemble technique} & 96.8 & 0.390\\
(\rom{4}) &\;(\rom{1}) {- Two-stage learning} & 120.8 & 0.376 \\
(\rom{5}) &\;(\rom{1}) {- SPE, deform conv} & 94.5 & 0.355\\
\bottomrule
\end{tabular}
\vskip 0.1in
\caption{\textbf{Quantitative evaluation on ablation study.}}
\label{tab:abl}
\end{table}

\section{Conclusion}
In this paper, we introduce an experimental protocol and a dedicated model for the panoramic Image-to-Image Translation (\ours) that considers 1) the structural properties of the panoramic images and 2) the lack of outdoor panoramic scene datasets.
To this end, we design our model to take panoramas as the source, and pinhole images with diverse conditions as the target, raising the large geometric variance between the source and target domains as a major challenging point.
To mitigate these issues, we propose distortion-aware panoramic modeling techniques and distortion-free discriminators to stabilize adversarial learning. Additionally, exploiting the cyclic property of panoramas, we propose to rotate and fuse the synthesized panoramas, resulting in the panorama output with a continuous view. 
We demonstrate the success of our method in translating realistic images in several benchmarks and look forward to future works that use our proposed experimental paradigm for panoramic image-to-image translation with non-pinhole camera inputs using diverse sets of pinhole image datasets.

\section*{Acknowledgements}
Most of this work was done while Soohyun Kim and Hwan Heo were research interns at NAVER AI Lab.
The NAVER Smart Machine Learning (NSML) platform~\cite{NSML} has been used in the experiments.

\clearpage
\newpage
{\small
\bibliographystyle{ieee_fullname}
\bibliography{egbib}
}

\appendix
\onecolumn

\clearpage

\newpage
\section*{Appendix}

\section{Implementation details}\label{supp:sec1}

\subsection{Architecture details}
We summarize the detailed network architecture of our method in \tabref{tab:1}.
We borrow the architecture of content encoder, style encoder, transformer blocks, and generator from InstaFormer\cite{kim2022instaformer} and discriminator from StarGANv2~\cite{choi2020stargan}. 'Layers for style encoder' inside the Encoder table indicates the end of the content encoder, while the style encoder has the same structure as the content encoder except for additional adaptive average pooling (AdaptiveAvgPool) and Conv-4, as shown below. Also, unlike the content encoder, the style encoder does not contain any normalization layer.
DeformConv indicates Deformable convolution with offset, and $(.)$ in the convolution indicates the zero-padding.

	\begin{table}[h] 
	\begin{center}
	 \scalebox{0.86}{
		\setlength{\tabcolsep}{10pt}
			\begin{tabular}{lcc}
			    \multicolumn{3}{c}{\textbf{Encoder}} \tabularnewline
				\midrule
				Layer & Parameters $(\mathtt{in},\mathtt{out},\mathtt{k},\mathtt{s},\mathtt{p})$ & Output shape $(C \times H \times W)$ \tabularnewline
				\midrule
				DeformConv-1 (Reflection) & $(3,64,7,1,3)$ & $(64,256,512)$  \tabularnewline
				InstanceNorm & - & $(64,256,512)$  \tabularnewline		
				ReLU & - & $(64,256,512)$  \tabularnewline		\midrule
				Conv-2 (Zeros) & $(64,128,3,1,1)$ & $(128,256,512)$ \tabularnewline
				InstanceNorm & - & $(128,256,512)$  \tabularnewline		
				ReLU & - & $(128,256,512)$  \tabularnewline		\midrule
				Downsample & - & $(128,128,256)$ \tabularnewline		
				Conv-3 (Zeros) & $(128,256,3,1,1)$ & $(256,128,256)$ \tabularnewline
				InstanceNorm & - & $(256,128,256)$  \tabularnewline		
				ReLU & - & $(256,128,256)$  \tabularnewline		
				DownSample & - & $(256,64,128)$ \tabularnewline \midrule
				\textbf{Layers for style encoder}& \tabularnewline
				AdaptiveAvgPool & - & $(256,1,1)$ \tabularnewline
				Conv-4 & $(256,8,1,1,0)$ & $(8,1,1)$ \tabularnewline
				\midrule
                &&\tabularnewline 
                \multicolumn{3}{c}{\textbf{Transformer Encoder}} \tabularnewline \midrule
				Layer & Parameters $(\mathtt{in},\mathtt{out})$ & Output shape $(C)$ \tabularnewline
				\midrule
				AdaptiveInstanceNorm & - & $(1024)$ \tabularnewline
                Linear-1 & $(1024,3072)$ & $(3072)$ \tabularnewline
				Attention & - & $(1024)$ \tabularnewline
				Linear-2 & $(1024,1024)$ & $(1024)$ \tabularnewline
				AdaptiveInstanceNorm & - & $(1024)$ \tabularnewline
                Linear-3 & $(1024,4096)$ & $(4096)$ \tabularnewline
				GELU & - & $(4096)$ \tabularnewline
				Linear-4 & $(4096,1024)$ & $(1024)$ \tabularnewline

				\midrule
                &&\tabularnewline 
                \multicolumn{3}{c}{\textbf{Generator}} \tabularnewline \midrule
				Layer & Parameters $(\mathtt{in},\mathtt{out},\mathtt{k},\mathtt{s},\mathtt{p})$ & Output shape $(C \times H \times W)$ \tabularnewline
				\midrule
                UpSample & - & $(256,128,256)$ \tabularnewline 
                Conv-1 (Zeros) & $(256,128,3,1,1)$ & $(128,128,256)$  \tabularnewline
				LayerNorm & - & $(128,128,256)$  \tabularnewline		
				ReLU & - & $(128,128,256)$  \tabularnewline		\midrule
                UpSample & - & $(128,256,512)$ \tabularnewline 
                Conv-2 (Zeros) & $(128,64,3,1,1)$ & $(64,256,512)$  \tabularnewline
				LayerNorm & - & $(64,256,512)$  \tabularnewline		
				ReLU & - & $(64,256,512)$  \tabularnewline		\midrule
				Conv-3 (ReflectionPad) & $(64,3,7,1,3)$ & $(3,256,512)$ \tabularnewline
				Tanh & - & $(3,256,512)$  \tabularnewline	
				\bottomrule
			\end{tabular}
 		}
            \vskip 0.1in
            \end{center}
		\caption{\textbf{Network architecture of Pano-I2I.}}
		\label{tab:1}
	\end{table}

\newpage
\subsection{Deformable convolution}
Our deformable convolution finds the fixed offset for ERP format, as in PAVER~\cite{yun2022panoramic} and PanoFormer~\cite{shen2022panoformer}.
As mentioned in the main paper, the deformable convolution layer is applied on the equirectangular (ERP) format of panorama image by deriving an ERP plane offset $\Theta_{\text{ERP}} \in \mathbb{R}^{2\times H \times W \times \text{ker}_h \times \text{ker}_w}$ for each kernel location (here, the kernel size is 7$\times$7) that considers the panoramic geometry. 
After obtaining the offset for once, we keep the kernel shape on the tangent fixed.
The conversion function from 3D-Cartesian domain to spherical domain and spherical domain to ERP domain:
\begin{align}
f_{\text{3D}\rightarrow\text{SPH}}(x,y,z) = (\text{arctan}\frac{y}{x}, \text{arctan}\frac{\sqrt{x^2 + y^2}}{z}),
f_{\text{SPH}\rightarrow\text{ERP}}(\theta, \phi) = (\frac{W}{2\pi}\phi, \frac{H}{\pi}\theta),
\end{align}
where W, H is width and height for the panoramic input, respectively, and $\theta\in[0,2\pi], \phi\in[0,\pi]$.

The rotation matrix is as follows:
\begin{align}
    R(\theta, \phi) &= 
    \begin{pmatrix}
    \cos\phi \cos\theta & -\cos\phi \sin\theta & \sin\theta \\
    \sin\theta & \cos\theta & 0 \\
    \sin\phi \cos\theta & -\sin\phi \sin\theta & \cos\phi
    \end{pmatrix}.
\end{align}

\subsection{Training details}
We employ the Adam optimizer, where $\beta_{1} = 0.9$ and $\beta_{2} = 0.99$, for 100 epochs using a step decay learning rate scheduler.
We also set a batch size of 8 for Stage \rom{1}, and 4 for Stage \rom{2} for each GPU. The initial learning rate is 1e-4. All coefficients for the losses are set to 1, except for $\lambda_\mathrm{NCE}^\mathrm{cont}$, which is set as 15, and $\lambda_\mathrm{df-GAN}$ is set as 0.8 for Stage \rom{2}.
The number of negative patches for content loss is 255.
The training images are resized to 256$\times$512. We conduct experiments using 8 Tesla V100 GPUs. The trained weights and code will be made publicly available.

\subsection{Notation}
We provide the notations that are used in the main paper, in \tabref{tab:notation}.

\begin{table}[ht!]
\begin{center}
\begin{tabular}{cl}
\toprule
Symbol & Definition \\
\midrule
$\mathbf{x}$  & Content image from source domain (panorama)\\
$\mathbf{y}$  & Style image from target domain (pinhole image)\\
$\mathbf{\hat{y}}$  & Translated image (panorama)\\
$\mathcal{E}_\mathrm{c}$  & Content encoder with Deformable Conv\\
$\mathcal{E}_\mathrm{s}$  & Style encoder with Deformable Conv\\
$\mathcal{T}$ & Transformer encoder\\
$\mathcal{G}$  & Generator\\
$\mathcal{D}$  & Discriminator\\
$\Theta$ & Offset used in deformable layer \\
$\theta$ & rotation angle in augmentation \\
$S$    & Length of one side of tangential square patches \\
$W, H$ & Sizes of 360$^\circ$ image input (\eg, $W=512, H=256$) \\
$w, h$ & Number of patches along width and height\\
$l$    & Number of channels per feature  \\
\bottomrule
\end{tabular}
\end{center}

\vspace{-5pt}
\caption{\textbf{Our notations are summarized.}}
\label{tab:notation}
\end{table}

\newpage
\subsection{Evaluation details}
\label{sec:eval}

\begin{figure}[h]
\begin{center}
\newcolumntype{M}[1]{>{\centering\arraybackslash}m{#1}}
\setlength{\tabcolsep}{1pt} 
\renewcommand{\arraystretch}{2} 
\footnotesize
\begin{tabular}
{M{0.015\linewidth}M{0.24\linewidth}M{0.001\linewidth}|M{0.001\linewidth}M{0.24\linewidth}M{0.001\linewidth}M{0.24\linewidth}M{0.001\linewidth}M{0.24\linewidth}}
 & Source &&& CUT~\cite{park2020contrastive} && FSeSim~\cite{zheng2021spatially} && \ours (ours) \\\hline 
& & & & & & &  \\ 
\vspace{-15pt}
\rotatebox{90}{Panorama} &
\vspace{-15pt}
{\includegraphics[width=\linewidth]{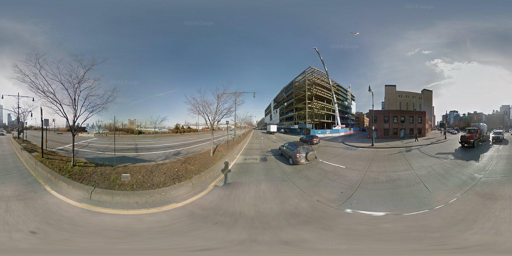}}\hfill  &&& 
\vspace{-15pt}
{\includegraphics[width=\linewidth]{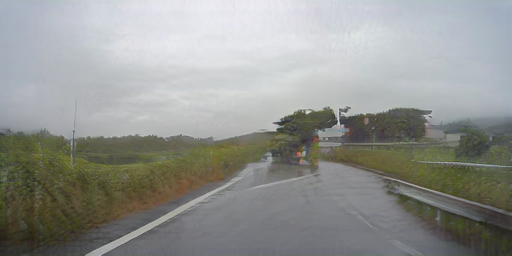}}\hfill && 
\vspace{-15pt}
{\includegraphics[width=\linewidth]{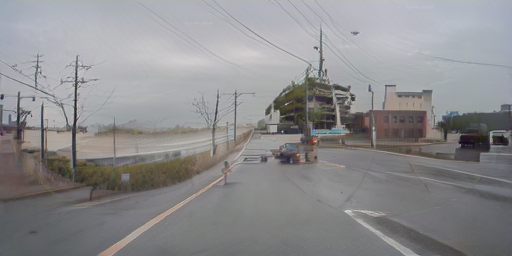}}\hfill  
&&
\vspace{-15pt}
{\includegraphics[width=\linewidth]{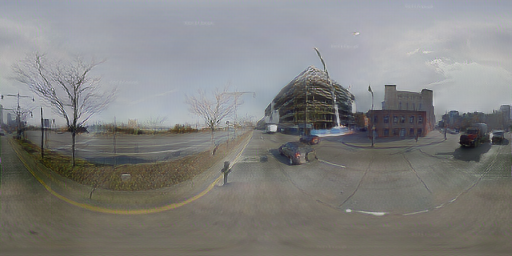}}\hfill \\
\hline 
& & & & & & & &  \\
\vspace{-15pt}
\multirow{2}{*}{\rotatebox{90}{Projected images}} &
\vspace{-15pt}{\includegraphics[
width=\linewidth]{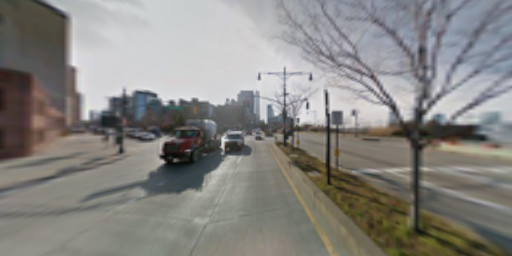}}\hfill
&&&
\vspace{-15pt}
{\includegraphics[width=\linewidth]{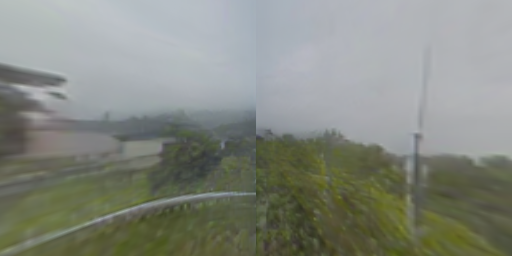}}\hfill
&&
\vspace{-15pt}
{\includegraphics[width=\linewidth]{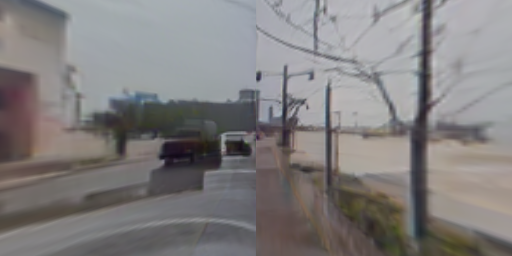}}\hfill
&&
\vspace{-15pt}
{\includegraphics[width=\linewidth]{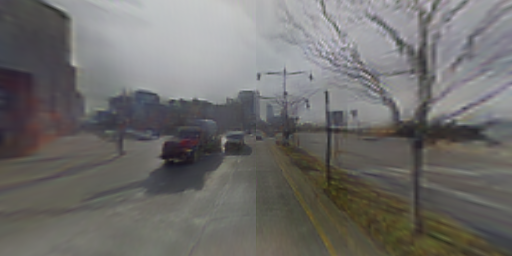}}\hfill\\

& & & & & & &  \\ 
& \vspace{-20pt}
{\includegraphics[width=\linewidth]{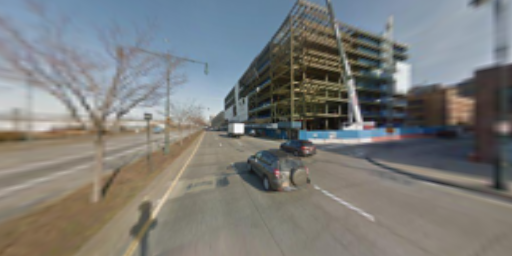}}\hfill
&&&
\vspace{-20pt}
{\includegraphics[width=\linewidth]{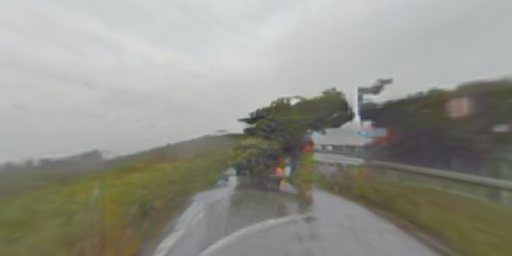}}\hfill
&&
\vspace{-20pt}
{\includegraphics[width=\linewidth]{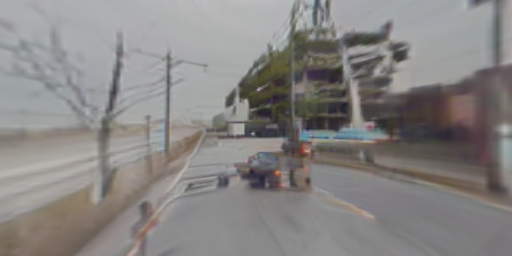}}\hfill
&&
\vspace{-20pt}
{\includegraphics[width=\linewidth]{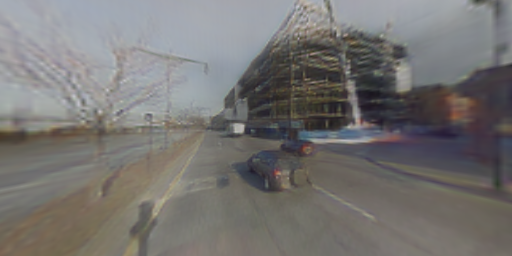}}\hfill\\

\end{tabular}\\
\end{center}
    \caption{\textbf{Visualization of panorama-to-pinhole image conversion.} We visualize the input and the output images (first row) and the examples of projected images (second row and third row) where $\phi$ is fixed as 0. Note that the projected source images are not used as input to the networks.}
    \label{fig:supp_prj}
\end{figure}

\paragrapht{Fr\'echet Inception Distance (FID)~\cite{heusel2017gans}} is computed by measuring the mean and variance distances between the generated and real images in the Inception feature space. We used the default setting of FID measurement provided in \footnote{ https://github.com/mseitzer/pytorch-fid}. In the main paper, we sampled 10 times for 1000 test images. Therefore, we computed the FID for each sampled set and averaged the scores to get the final result, and evaluated FID between target images and output images to measure style relevance.
In addition, considering that the structure of outputs tends to become pinhole-like in panoramic I2I tasks, we measure the FID metric after applying panorama-to-pinhole projection ($f_T$) for randomly chosen horizontal angle $\theta$ and fixed vertical angle $\phi$ as 0 with a fixed FoV of \ang{90} to the output images, for consistent viewpoint with the target images. We visualize some examples of the projected images, compared with other methods in \figref{fig:supp_prj}. It should be noted that we measure FID between the original target images and the projected output images.

However, we noticed that the synthesized images of existing methods seem to follow not only the style of target images but also the pinhole structure and its contents (\textit{e.g.}, appearance of road, buildings, cars). In this regard, the higher FID score for style relevance does not guarantee better stylization results in this task.
Therefore, we additionally adopt the FID metric to measure structural distributions between source images and output images. To exclude the style information, we conduct such measurement in grayscale image format, shown in \tabref{tab:quan1}. We indicate such FID measurement as FID$_c$.

\paragrapht{Structural Similarity Index Measure (SSIM)~\cite{wang2004image}} is a widely used full-reference image quality assessment (IQA) measure, which measures the similarity between two images, where one of them is the reference image. We adopt the SSIM metric to measure the structural similarity between the source image and the output image.

\newpage
\section{More quantitative results}\label{supp:sec2}
In this section, we report additional quantitative results in \tabref{tab:quan1} to complement the main results.
As explained in \secref{sec:eval}, we additionally adopt the FID$_c$ metric to measure structural distributions between source panoramas and output panoramas, unlike the FID in the main paper. To exclude the style information, we transform images into a grayscale format.

\begin{table*}[h]
\begin{center}

\setlength{\tabcolsep}{8pt}
\begin{tabular}{lcc}
\toprule
\multirow{2}{*}{Method} & 
{Day$\rightarrow$Night} &  {Day$\rightarrow$Rainy}\\
\cmidrule(lr){2-3} 
& FID$_c\downarrow$ &  FID$_c\downarrow$\\ \midrule
CUT~\cite{park2020contrastive}      & 225.60  & 153.72   \\
FSeSim~\cite{zheng2021spatially}   & 179.28  & 136.44 \\
MGUIT~\cite{jeong2021memory}         & 433.17 & 147.38  \\
InstaFormer~\cite{kim2022instaformer}   & 231.38  & 149.91 \\
\midrule
\ours (ours) & \textbf{85.13} & \textbf{85.49}  \\
\bottomrule
\end{tabular} 
\end{center}
\caption{\textbf{Quantitative evaluation} in terms of 
the matching of feature distributions (FID~\cite{heusel2017gans} metric) on StreetLearn dataset~\cite{mirowski2018learning} to INIT dataset~\cite{shen2019towards}. Note that the purpose of FID measurement is different from the main paper; here we additionally adopt FID metric to measure structural distance.
}
\label{tab:quan1}
\end{table*}

\section{Additional results on ablation study}\label{supp:sec3}
In the main paper, we have examined the impacts of distortion-free discrimination, rotational ensemble, SPE and deformable convolution, and two-stage learning with quantitative and qualitative results in \tabref{tab:abl} and \figref{fig:ablation}. In this section, we provide additional visual results on day$\rightarrow$night (StreetLearn~\cite{mirowski2018learning} to INIT~\cite{shen2019towards}). 

\begin{figure*}[h]
\begin{center}
  \renewcommand{\thesubfigure}{}
    \subfigure[]{\includegraphics[width= 0.33\linewidth]{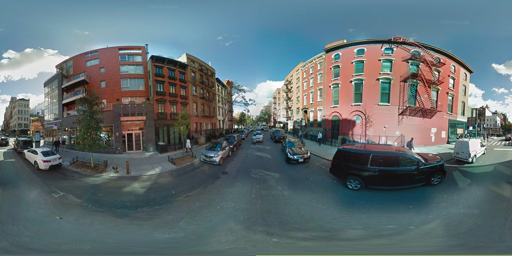}}\hfill
  \subfigure[]
{\includegraphics[width=  0.33\linewidth]{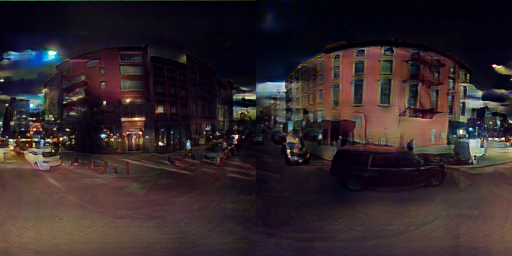}}\hfill
 \subfigure[]
{\includegraphics[width=  0.33\linewidth]{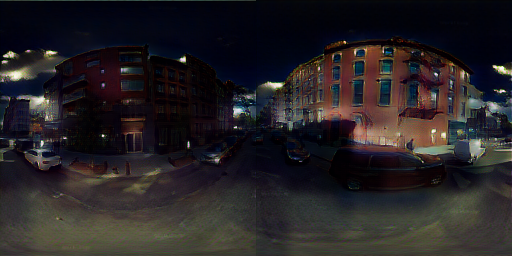}}\hfill\\ \vspace{-20pt}
   \subfigure[]
{\includegraphics[width=  0.33\linewidth]{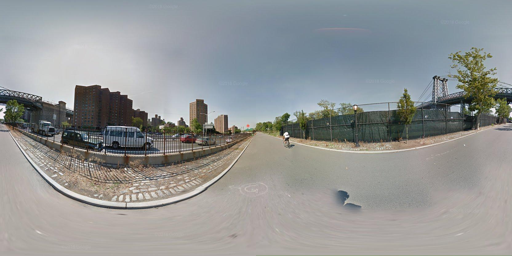}}\hfill
  \subfigure[]
{\includegraphics[width=  0.33\linewidth]{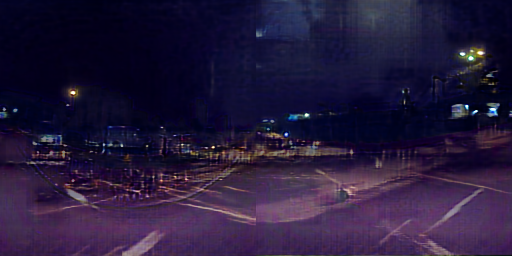}}\hfill
  \subfigure[]
{\includegraphics[width=  0.33\linewidth]{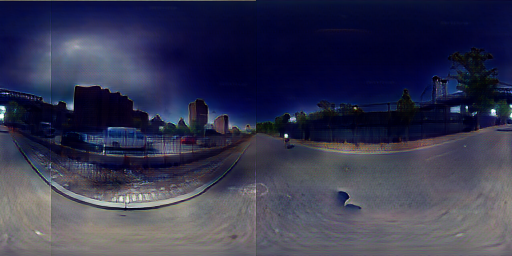}}\hfill\\\vspace{-20pt}
   \subfigure[Input]
{\includegraphics[width=  0.33\linewidth]{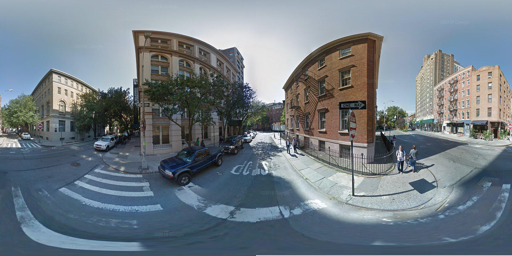}}\hfill
  \subfigure[- Distortion-free $\mathcal{D}$]
{\includegraphics[width=  0.33\linewidth]{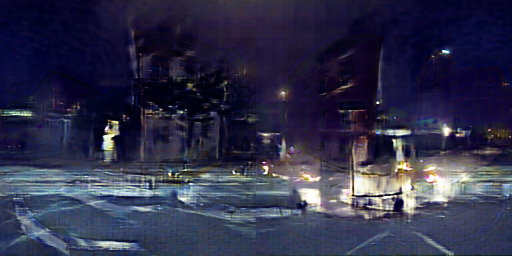}}\hfill 
  \subfigure[- Ensemble]
{\includegraphics[width=  0.33\linewidth]{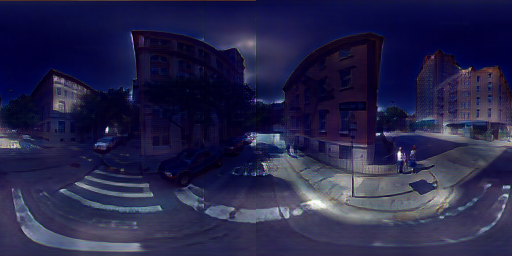}}\hfill\\\vspace{-5pt}
  \subfigure[]
{\includegraphics[width=  0.33\linewidth]{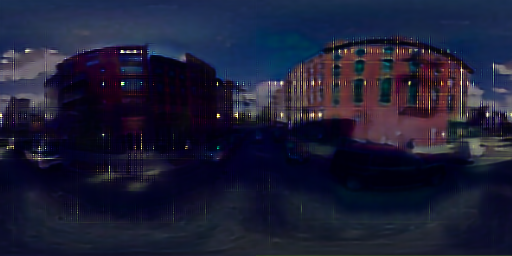}}\hfill
   \subfigure[]
{\includegraphics[width=  0.33\linewidth]{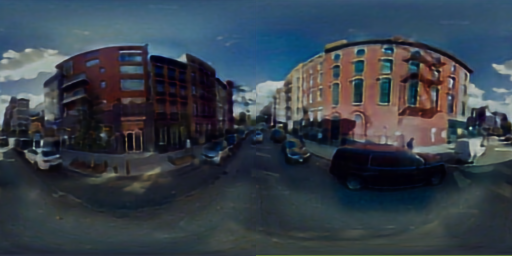}}\hfill
  \subfigure[]
{\includegraphics[width=  0.33\linewidth]{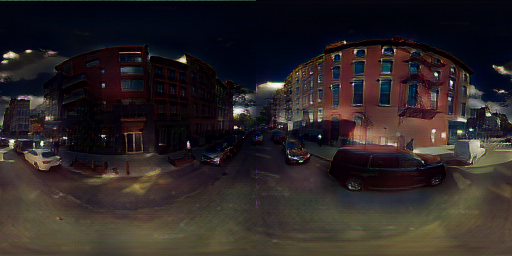}}\hfill\\\vspace{-20pt}
  \subfigure[]
{\includegraphics[width=  0.33\linewidth]{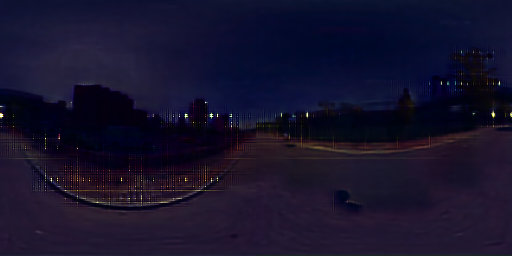}}\hfill
   \subfigure[]
{\includegraphics[width=  0.33\linewidth]{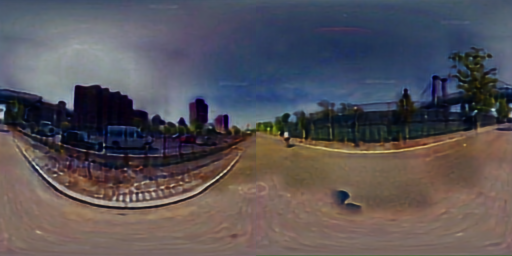}}\hfill
  \subfigure[]
{\includegraphics[width=  0.33\linewidth]{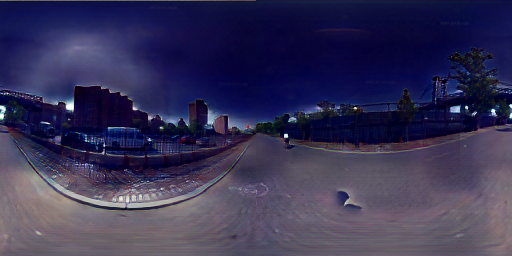}}\hfill\\\vspace{-20pt}
  \subfigure[- Two-stage learning]
{\includegraphics[width=  0.33\linewidth]{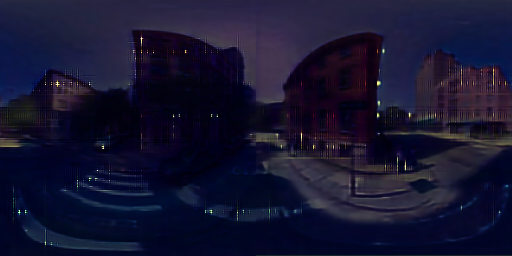}}\hfill
 \subfigure[- SPE, deform conv]
{\includegraphics[width=  0.33\linewidth]{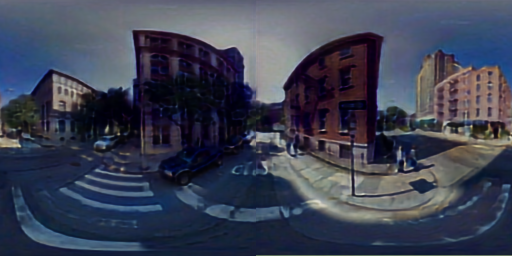}}\hfill
  \subfigure[\ours (ours)]
{\includegraphics[width=  0.33\linewidth]{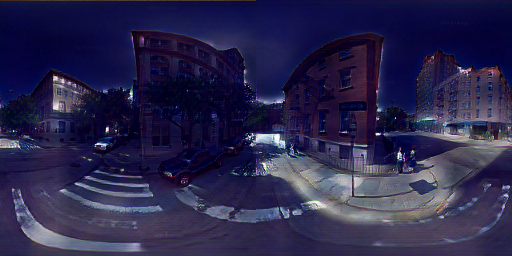}}\hfill\\
  \vspace{-5pt}
\end{center}
\caption{\textbf{Additional qualitative evaluation on ablation study.}}
  \label{fig:supp_abl}
\end{figure*}

As seen in \figref{fig:supp_abl}, our full model can maintain the boundary with high-quality generation, successfully preserving the panoramic structure.
It should be noted that the visual results are $\theta$=\ang{180} rotated to highlight the continuity at the edges. 
Especially, we can observe the ability of our discrimination design to generate distortion-tolerate outputs, and the results without ensemble technique fail to represent consistent style within an image. Also, the results without SPE and deformable convolution show a limited capability to capture structural continuity.

\clearpage
\newpage
\section{Training algorithms}\label{supp:sec4}
Here we provide the training algorithms for our stage-wise learning in pseudo-code forms. The first stage aims to reconstruct input panoramas, and the second stage learns to translate panoramas to have the style of pinhole images while preserving panoramic structure.

\subsection{Training algorithm for stage-\rom{1}}

\RestyleAlgo{ruled}
\begin{algorithm}[hbt!]
\caption{Pseudo code for stage-\rom{1}}\label{alg:one}
\SetKwInOut{Input}{Inputs}
\SetKwInOut{Output}{Output}
\Input{
\par
\begin{tabular}{l l l}
$\mathcal{X}$ & Source domain (panorama) \\
$\mathcal{E}_\mathrm{c}, \mathcal{E}_\mathrm{s}$  & Content and style encoders with a deformable convolution layer\\
$\mathcal{T}$ & Transformer encoder\\
$\mathcal{G}$  & Generator\\
$\mathcal{D}$  & Discriminator\\
$\xi$ & Initial parameters for $\mathcal{E}_\mathrm{c}$, $\mathcal{E}_\mathrm{s}$, $\mathcal{T}$, $\mathcal{G}$, and $\mathcal{D}$\\
$\Theta_{\mathrm{ERP}}$ & Equirectangular plane offset used in the deformable convolution layer \\
$\theta$ & Rotation angle in panoramic augmentation \\
$S_{l}$ & Number of patches in each $l$-th layer\\
$N$    & Total number of optimization steps\\
$\eta_n$ & Learning rate for $n$-th optimization step\\
$\alpha$ & Panoramic ensemble ratio \\
$\gamma(\cdot)$ & Sinusoidal mapping s.t. $\gamma(a) =\{ 
    (\mathrm{sin}(2^{k-1}\pi \textit{a}),\mathrm{cos}(2^{k-1}\pi \textit{a})) 
    | k=1,...,K\}$\\
$\ell(\cdot)$ & InfoNCE loss with a positive pair $(\mathbf{v}, \mathbf{v}^{+})$, negative pairs $(\mathbf{v}, \mathbf{v}^{-})$, and a temperature $\tau$ s.t. \\
    & $\ell(\hat{\mathbf{v}}, \mathbf{v}^{+}, \mathbf{v}^{-})=\mathrm{-log}\left[
    \frac{\mathrm{exp}({\mathbf{v}}\cdot \mathbf{v}^{+}/\tau)}{
    \mathrm{exp}({\mathbf{v}}\cdot \mathbf{v}^{+}/\tau) + \sum\mathrm{exp}({\mathbf{v}}\cdot \mathbf{v}^{-}/\tau)}\right]$\\
\end{tabular}
}

\For{$n=1$ \KwTo $N$}{
    $\mathcal{B} \gets \{\mathrm{x}|\mathrm{x}\sim\mathcal{X}\}$ \\
    \For{$(\mathrm{x})\in\mathcal{B}$}{
        $\mathrm{x}' \gets \mathrm{PanoramicAugment}(\mathrm{x}, \theta)$\\
        $\mathbf{c}^\mathrm{x}, \mathbf{c}^\mathrm{x'} \gets \mathcal{E}_\mathrm{c}(\mathrm{x}, \Theta_\mathrm{ERP}),  \mathcal{E}_\mathrm{c}(\mathrm{x'}, \Theta_\mathrm{ERP})$\\
        $\mathbf{s}^\mathrm{x} \gets \mathcal{E}_\mathrm{s}(\mathrm{x}, \Theta_\mathrm{ERP})$\\

        $\mathbf{\hat{y}^{(0)}}, \mathbf{\hat{y}^{(1)}} \gets \mathcal{G(\mathcal{T}(\mathbf{c}^\mathrm{x},\mathbf{s}^\mathrm{x}))},
        \mathcal{G(\mathcal{T}(\mathbf{c}^\mathrm{x'},\mathbf{s}^\mathrm{x}))}$ \\
        
        $\mathbf{\hat{y}} \gets \alpha \cdot \mathbf{\hat{y}^{(0)}} +(1-\alpha) \cdot  \mathrm{PanoramicAugment}({\mathbf{{\hat{y}^{(1)}}}}, -\theta)$\\
        
        $\mathbf{c}^{\mathrm{x}}, \mathbf{c}^{\hat{\mathrm{y}}} \gets \mathcal{E}_\mathrm{c}(\mathrm{\mathrm{x}}, \Theta_\mathrm{ERP}), \mathcal{E}_\mathrm{c}(\mathrm{\hat{\mathrm{y}}}, \Theta_\mathrm{ERP})$\\
        
        $\hat{\mathbf{v}}, \mathbf{v}^{+}, \mathbf{v}^{-} \gets \mathrm{PatchSample}_\mathrm{pos}(\mathbf{c}^{\hat{\mathrm{y}}}), \mathrm{PatchSample}_\mathrm{pos}(\mathbf{c}^{\mathrm{x}}), \mathrm{PatchSample}_\mathrm{neg}(\mathbf{c}^{\mathrm{x}})$\\

        $\mathbf{s}^{\hat{\mathrm{y}}} \gets \mathcal{E}_\mathrm{s}(\mathrm{\hat{\mathrm{y}}}, \Theta_\mathrm{ERP})$\\
        
        $\mathcal{L}_{\mathrm{GAN}} \gets \mathrm{log}(1-\mathcal{D}(\hat{\mathbf{y}}))+ \mathrm{log}\, \mathcal{D}(\mathbf{x})$\\
        
        $\mathcal{L}_\mathrm{NCE}^{\mathrm{cont}} \gets \sum\limits_{l}\sum\limits_{s}\ell(\hat{\mathbf{v}}, \mathbf{v}^{+}, \mathbf{v}^{-})$\\
        
        $\mathcal{L}_{\mathrm{ref}\minus \mathrm{recon}}^\mathrm{style} \gets \|
    {\mathbf{s}^{\hat{\mathrm{y}}} - \mathbf{s}^\mathrm{x}}\|_{1}$
    }
    $\delta\xi \gets \frac{1}{N} \sum\limits_{(\mathrm{x}, \mathrm{y}) \in \mathcal{B}}  \mathcal{L}_{\mathrm{GAN}}
+\lambda_{\mathrm{cont}}\mathcal{L}_\mathrm{NCE}^{\mathrm{cont}} 
+\lambda_\mathrm{style}\mathcal{L}_{\mathrm{ref}\minus \mathrm{recon}}^\mathrm{style}$\\
    $\xi \gets \mathrm{optimizer}(\xi, \delta\xi, \eta_n)$
}

\end{algorithm}

\newpage
\subsection{Training algorithm for stage-\rom{2}}

\RestyleAlgo{ruled}
\begin{algorithm}[hbt!]
\caption{Pseudo code for stage-\rom{2}}\label{alg:two}
\SetKwInOut{Input}{Inputs}
\SetKwInOut{Output}{Output}
\Input{
\par
\begin{tabular}{l l l}
$\mathcal{X}$ & Source domain (panorama) \\
$\mathcal{Y}$ & Target domain (pinhole image) \\
$\mathcal{E}_\mathrm{c}, \mathcal{E}_\mathrm{s}$  & Content and style encoders with a deformable convolution layer\\
$\mathcal{T}$ & Transformer encoder\\
$\mathcal{G}$  & Generator\\
$\mathcal{D}$  & Discriminator\\
$f_T$ & Panorama-to-pinhole image conversion function \\
$\xi$ & Initial parameters for $\mathcal{E}_\mathrm{c}$, $\mathcal{E}_\mathrm{s}$, $\mathcal{T}$, $\mathcal{G}$, $\mathcal{D}$, and $f_T$\\
$\Theta_{\mathrm{ERP}}$ & Equirectangular plane offset used in the deformable convolution layer \\
$\Theta_{\varnothing}$ & Zero offset \\
$\theta$ & Rotation angle in panoramic augmentation \\
$S_{l}$ & Number of patches in each $l$-th layer\\
$N$    & Total number of optimization steps\\
$\eta_n$ & Learning rate for $n$-th optimization step\\
$\alpha$ & Panoramic ensemble ratio \\
$\gamma(\cdot)$ & Sinusoidal mapping s.t. $\gamma(a) =\{ 
    (\mathrm{sin}(2^{k-1}\pi \textit{a}),\mathrm{cos}(2^{k-1}\pi \textit{a})) 
    | k=1,...,K\}$\\
$\ell(\cdot)$ & InfoNCE loss with a positive pair $(\mathbf{v}, \mathbf{v}^{+})$, negative pairs $(\mathbf{v}, \mathbf{v}^{-})$, and a temperature $\tau$ s.t. \\
    & $\ell(\hat{\mathbf{v}}, \mathbf{v}^{+}, \mathbf{v}^{-})=\mathrm{-log}\left[
    \frac{\mathrm{exp}({\mathbf{v}}\cdot \mathbf{v}^{+}/\tau)}{
    \mathrm{exp}({\mathbf{v}}\cdot \mathbf{v}^{+}/\tau) + \sum\mathrm{exp}({\mathbf{v}}\cdot \mathbf{v}^{-}/\tau)}\right]$\\
\end{tabular}
}

\For{$n=1$ \KwTo $N$}{
    $\mathcal{B} \gets \{(\mathrm{x}, \mathrm{y})|\mathrm{x}\sim\mathcal{X}, \mathrm{y}\sim\mathcal{Y}\}$ \\
    \For{$(\mathrm{x}, \mathrm{y})\in\mathcal{B}$}{
        $\mathrm{x}' \gets \mathrm{PanoramicAugment}(\mathrm{x}, \theta)$\\
        $\mathbf{c}^\mathrm{x}, \mathbf{c}^\mathrm{x'} \gets \mathcal{E}_\mathrm{c}(\mathrm{x}, \Theta_\mathrm{ERP}),  \mathcal{E}_\mathrm{c}(\mathrm{x'}, \Theta_\mathrm{ERP})$\\
        $\mathbf{s} \sim \mathcal{N}(0, \mathbf{I})$\\

        $\mathbf{\hat{y}^{(0)}}, \mathbf{\hat{y}^{(1)}} \gets \mathcal{G(\mathcal{T}(\mathbf{c}^\mathrm{x},\mathbf{s}))},
        \mathcal{G(\mathcal{T}(\mathbf{c}^\mathrm{x'},\mathbf{s}))}$ \\
        
        $\mathbf{\hat{y}} \gets \alpha \cdot \mathbf{\hat{y}^{(0)}} +(1-\alpha) \cdot  \mathrm{PanoramicAugment}({\mathbf{{\hat{y}^{(1)}}}}, -\theta)$\\
        
        $\mathbf{c}^{\mathrm{x}}, \mathbf{c}^{\mathrm{y}}, \mathbf{c}^{\hat{\mathrm{y}}} \gets 
        \mathcal{E}_\mathrm{c}(\mathrm{\mathrm{x}}, \Theta_\mathrm{ERP}),
        \mathcal{E}_\mathrm{c}(\mathrm{\mathrm{y}}, \Theta_{\varnothing}), \mathcal{E}_\mathrm{c}(\mathrm{\hat{\mathrm{y}}}, \Theta_\mathrm{ERP})$\\
        
        $\hat{\mathbf{v}}, \mathbf{v}^{+}, \mathbf{v}^{-} \gets \mathrm{PatchSample}_\mathrm{pos}(\mathbf{c}^{\hat{\mathrm{y}}}), \mathrm{PatchSample}_\mathrm{pos}(\mathbf{c}^{\mathrm{x}}), \mathrm{PatchSample}_\mathrm{neg}(\mathbf{c}^{\mathrm{x}})$\\

        $\mathbf{s}^{\mathrm{y}}, \mathbf{s}^{\hat{\mathrm{y}}} \gets \mathcal{E}_\mathrm{s}(\mathrm{\hat{\mathrm{y}}}, \Theta_{\varnothing}), \mathcal{E}_\mathrm{s}(\mathrm{\hat{\mathrm{y}}}, \Theta_\mathrm{ERP})$\\
        
        $\mathcal{L}_{\mathrm{GAN}} \gets \mathrm{log}(1-\mathcal{D}(\hat{\mathbf{y}}))+ \mathrm{log}\, \mathcal{D}(\mathbf{x})$\\
        $\mathcal{L}_{\mathrm{df}\minus\mathrm{GAN}} \gets \mathrm{log}(1-\mathcal{D}(f_T(\hat{\mathbf{y}})))+ \mathrm{log}\, \mathcal{D}(\mathbf{y})$\\
        
        $\mathcal{L}_\mathrm{NCE,\mathrm{x}}^{\mathrm{cont}} \gets \sum\limits_{l}\sum\limits_{s}\ell(\hat{\mathbf{v}}, \mathbf{v}^{+}, \mathbf{v}^{-})$\\
        
        $\mathcal{L}_{\mathrm{rand}\minus \mathrm{recon}}^\mathrm{style} \gets \|
    {\mathbf{s}^{\hat{\mathrm{y}}} - \mathbf{s}}\|_{1}$
    
        $\mathcal{L}_{\mathrm{recon}}^\mathrm{img} \gets\|{\mathcal{G}(\mathcal{T}(\contenty,\styley)) -\,\mathbf{y}}\|_{1}$\\
    }
    $\delta\xi \gets \frac{1}{N} \sum\limits_{(\mathrm{x}, \mathrm{y}) \in \mathcal{B}} \partial_\xi \mathcal{L}_{\mathrm{df}\minus\mathrm{GAN}} + (1-\partial_\xi)\mathcal{L}_{\mathrm{GAN}}
+\lambda_{\mathrm{cont}}\mathcal{L}_\mathrm{NCE}^{\mathrm{cont}} 
+\lambda_\mathrm{style}\mathcal{L}_{\mathrm{rand}\minus \mathrm{recon}}^\mathrm{style} + \lambda_\mathrm{recon}\mathcal{L}_{\mathrm{recon}}^\mathrm{img}$ \\
    $\xi \gets \mathrm{optimizer}(\xi, \delta\xi, \eta_n)$
}

\end{algorithm}

\clearpage
\newpage
\section{Additional qualitative results}\label{supp:sec5}

\subsection{INIT dataset}
\begin{figure*}[h]
\centering
\renewcommand{\thesubfigure}{}
    \subfigure[]{\includegraphics[width= 0.33\linewidth]
{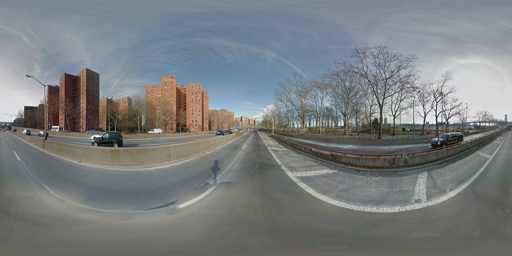}}
    \hfill
    \subfigure[]{\includegraphics[width= 0.33\linewidth]{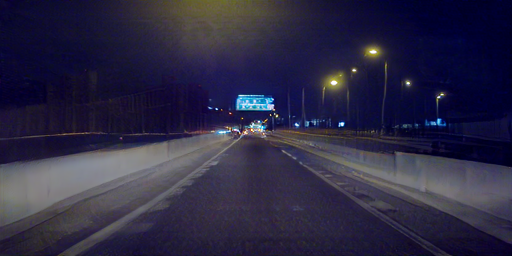}}
    \hfill
     \subfigure[]{\includegraphics[width= 0.33\linewidth]{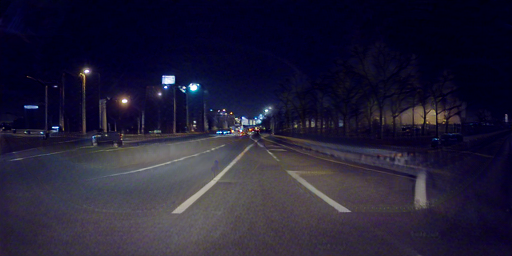}}
    \hfill \\ \vspace{-20pt}
    
    \subfigure[]{\includegraphics[width= 0.33\linewidth]{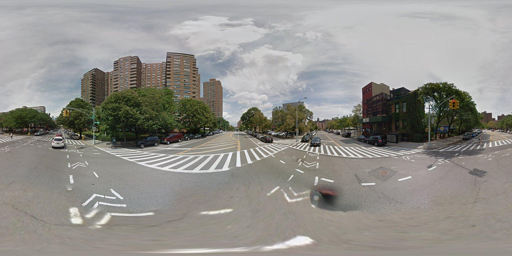}}
    \hfill
    \subfigure[]{\includegraphics[width= 0.33\linewidth]{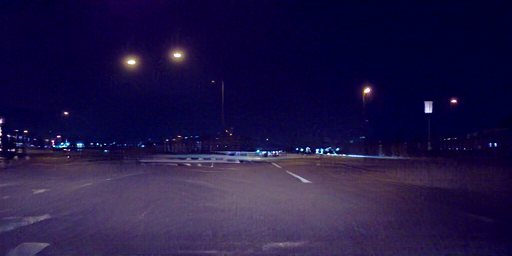}}
    \hfill
    \subfigure[]{\includegraphics[width= 0.33\linewidth]{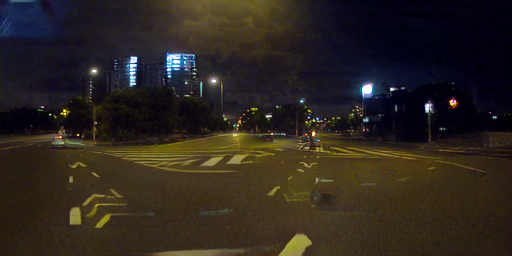}}
    \hfill \\ \vspace{-20pt}
    
    \subfigure[Inputs]{\includegraphics[width= 0.33\linewidth]{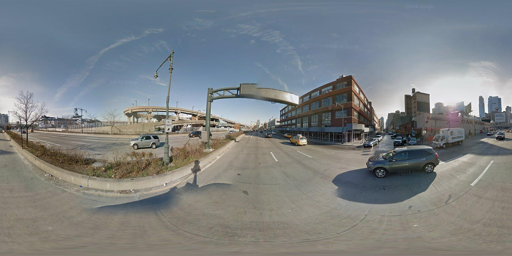}}
    \hfill
    \subfigure[CUT~\cite{park2020contrastive}]{\includegraphics[width= 0.33\linewidth]{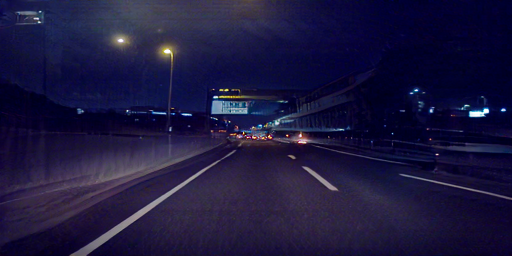}}
    \hfill
    \subfigure[FSeSim~\cite{zheng2021spatially}]{\includegraphics[width= 0.33\linewidth]{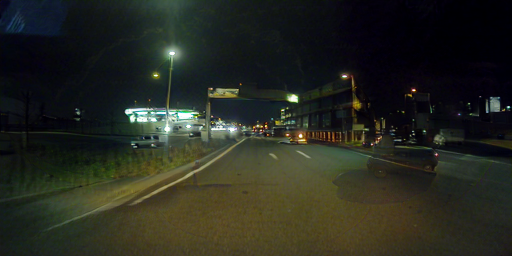}}
    \hfill \\ \vspace{-5pt}
    
    \subfigure[]{\includegraphics[width= 0.33\linewidth]{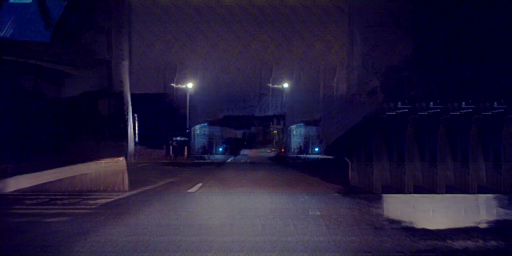}}
    \hfill
    \subfigure[]{\includegraphics[width= 0.33\linewidth]{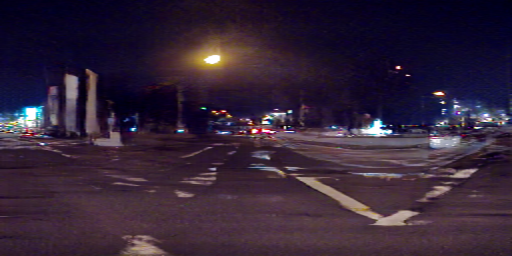}}
    \hfill
    \subfigure[]{\includegraphics[width= 0.33\linewidth]{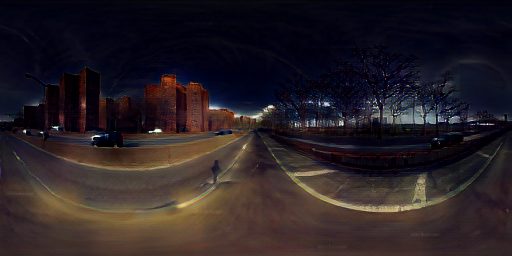}}
    \hfill\\ \vspace{-20pt}

    \subfigure[]{\includegraphics[width= 0.33\linewidth]{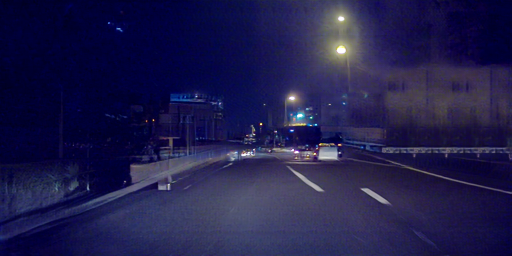}}
    \hfill
    \subfigure[]{\includegraphics[width= 0.33\linewidth]{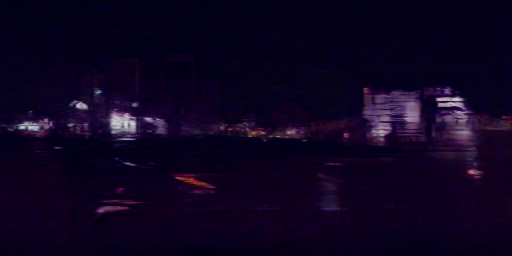}}
    \hfill
    \subfigure[]{\includegraphics[width= 0.33\linewidth]{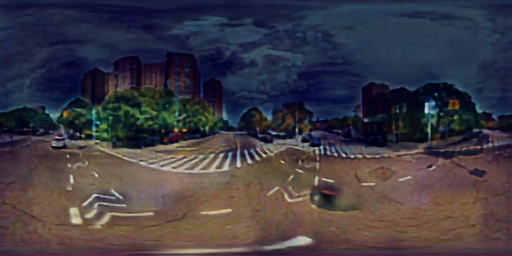}}
    \hfill\\ \vspace{-20pt}
    \subfigure[InstaFormer~\cite{kim2022instaformer}]{\includegraphics[width= 0.33\linewidth]{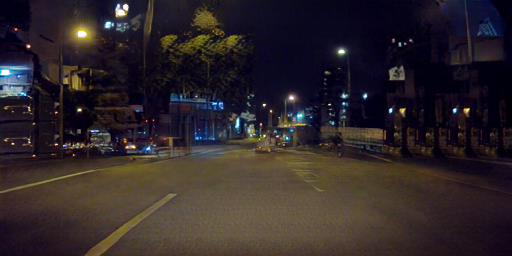}}
    \hfill
    \subfigure[MGUIT~\cite{jeong2021memory}]{\includegraphics[width= 0.33\linewidth]{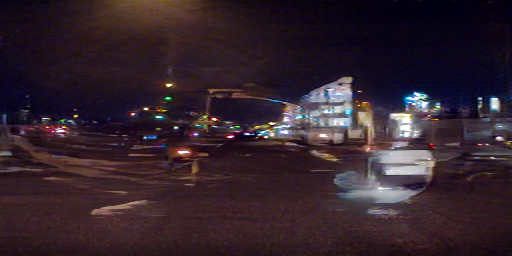}}
    \hfill
    \subfigure[\ours (ours)]{\includegraphics[width= 0.33\linewidth]{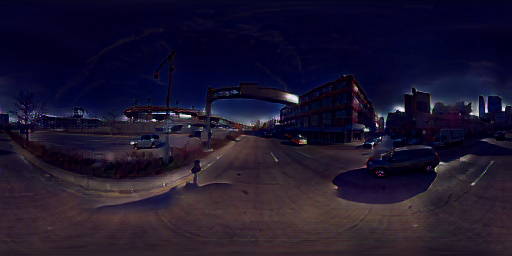}}
    \hfill\\
    
    \caption{\textbf{Qualitative comparison} on StreetLearn dataset~\cite{mirowski2018learning} (day) to INIT dataset~\cite{shen2019towards} (night).}
    \label{fig:abl_init_night}
\end{figure*}

\begin{figure*}[h]
\centering

    \renewcommand{\thesubfigure}{}
    \subfigure[]{\includegraphics[width= 0.33\linewidth]{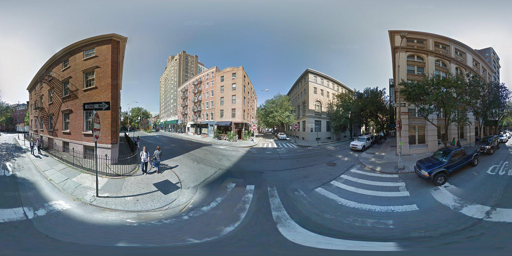}}
    \hfill
    \subfigure[]{\includegraphics[width= 0.33\linewidth]{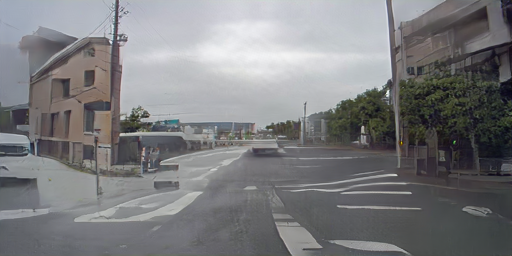}}
    \hfill
    \subfigure[]{\includegraphics[width= 0.33\linewidth]{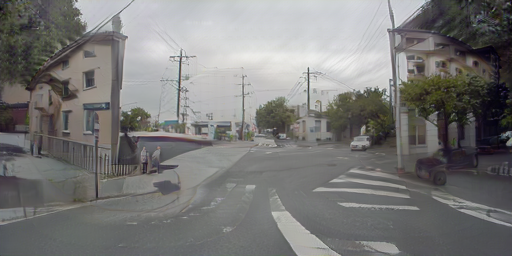}}
    \hfill \\ \vspace{-20pt}
    
    \subfigure[]{\includegraphics[width= 0.33\linewidth]{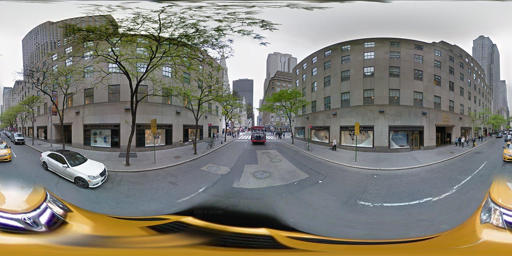}}
    \hfill
    \subfigure[]{\includegraphics[width= 0.33\linewidth]{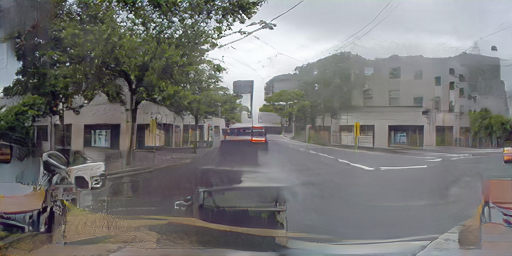}}
    \hfill
    \subfigure[]{\includegraphics[width= 0.33\linewidth]{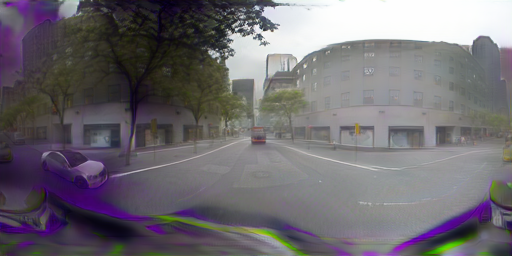}}
    \hfill \\ \vspace{-20pt}
    
    \subfigure[Inputs]{\includegraphics[width= 0.33\linewidth]{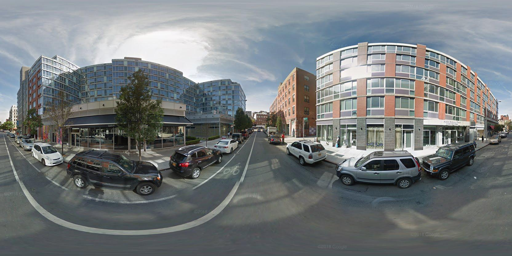}}
    \hfill
    \subfigure[CUT~\cite{park2020contrastive}]{\includegraphics[width= 0.33\linewidth]{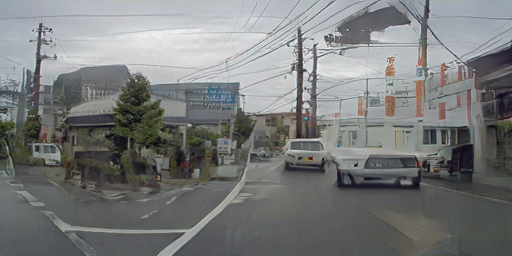}}
    \hfill
    \subfigure[FSeSim~\cite{zheng2021spatially}]{\includegraphics[width= 0.33\linewidth]{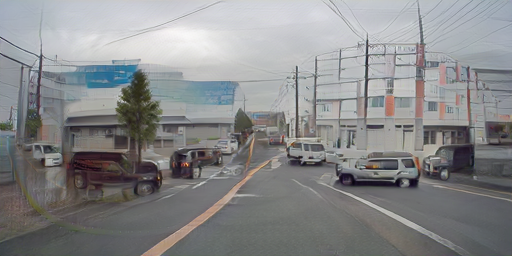}}
    \hfill \\

    \subfigure[]{\includegraphics[width= 0.33\linewidth]{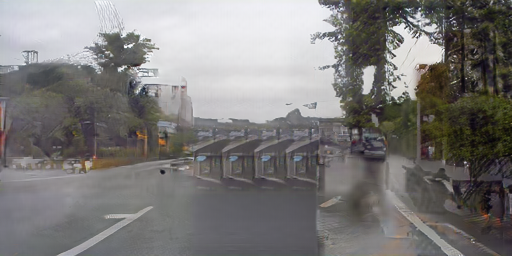}}
    \hfill
    \subfigure[]{\includegraphics[width= 0.33\linewidth]{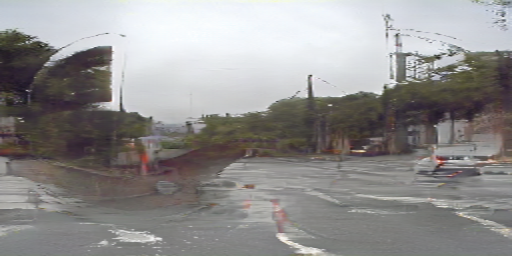}}
    \hfill
    \subfigure[]{\includegraphics[width= 0.33\linewidth]{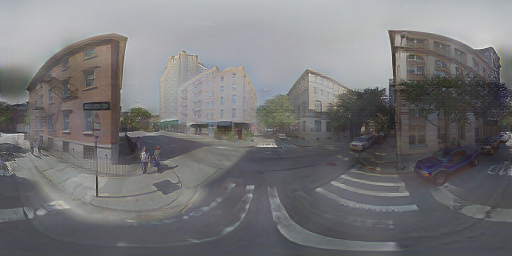}}
    \hfill \\ \vspace{-20pt}
    
    \subfigure[]{\includegraphics[width= 0.33\linewidth]{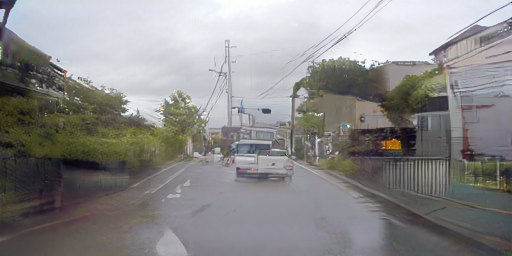}}
    \hfill
    \subfigure[]{\includegraphics[width= 0.33\linewidth]{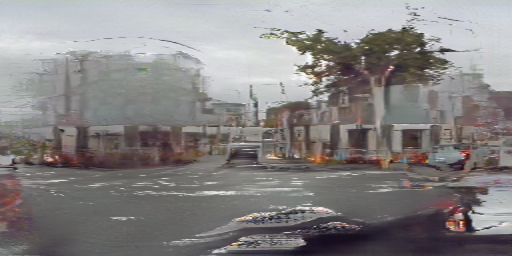}}
    \hfill
    \subfigure[]{\includegraphics[width= 0.33\linewidth]{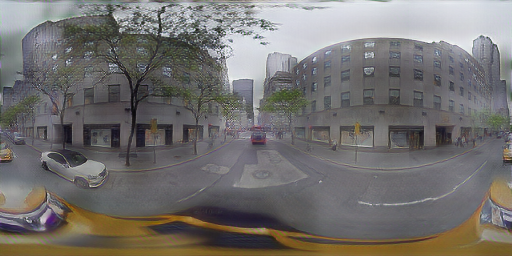}}
    \hfill \\ \vspace{-20pt}
    
    \subfigure[InstaFormer~\cite{kim2022instaformer}]{\includegraphics[width= 0.33\linewidth]{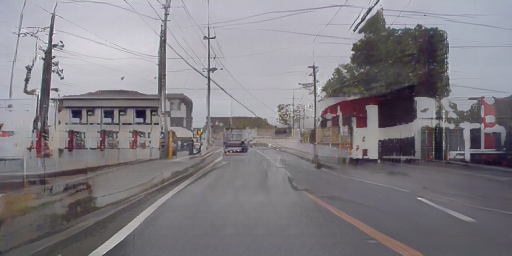}}
    \hfill
    \subfigure[MGUIT~\cite{jeong2021memory}]{\includegraphics[width= 0.33\linewidth]{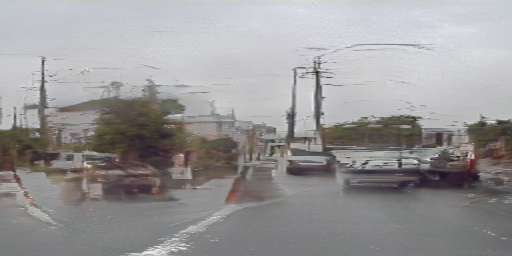}}
    \hfill
    \subfigure[\ours (ours)]{\includegraphics[width= 0.33\linewidth]{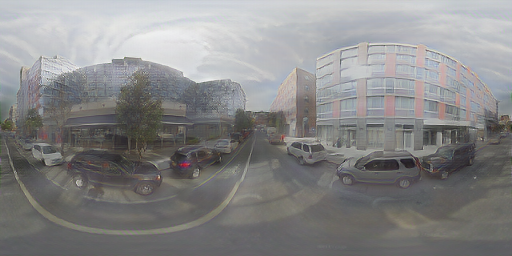}}
    \hfill \\

    \caption{\textbf{Qualitative comparison} on StreetLearn dataset~\cite{mirowski2018learning} (day) to INIT dataset~\cite{shen2019towards} (rainy).}
    \label{fig:abl_init_rainy}
\end{figure*}

We include additional qualitative comparisons to other I2I methods, CUT~\cite{park2020contrastive}, FSeSim~\cite{zheng2021spatially} MGUIT~\cite{jeong2021memory}, and Instaformer~\cite{kim2022instaformer} on StreetLearn to various conditions of INIT: day$\rightarrow$night in \figref{fig:abl_init_night}, day$\rightarrow$rainy in \figref{fig:abl_init_rainy}.

\clearpage
\newpage
\subsection{Dark Zurich dataset}
\begin{figure*}[h]
  \begin{center}
\renewcommand{\thesubfigure}{}
    \subfigure[]{\includegraphics[width= 0.33\linewidth]{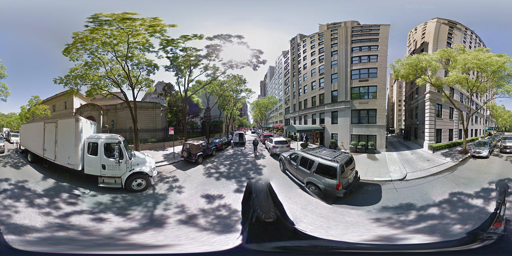}}
    \hfill
    \subfigure[]{\includegraphics[width= 0.33\linewidth]{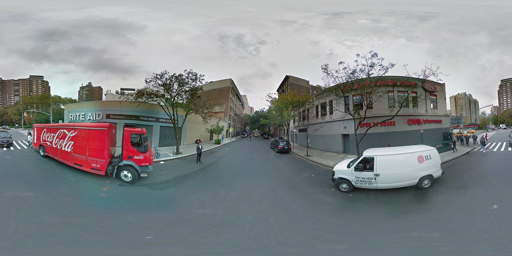}}
    \hfill
    \subfigure[]{\includegraphics[width= 0.33\linewidth]{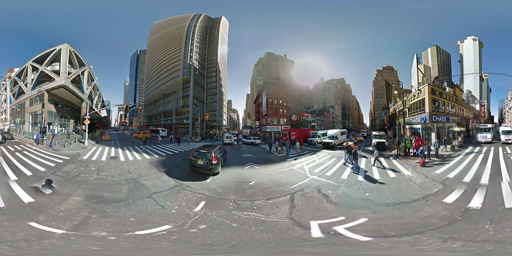}}
    \hfill \\\vspace{-10pt}
    
    {\small (a) Inputs} \\ \vspace{3pt}
  
    \subfigure[]{\includegraphics[width= 0.33\linewidth]{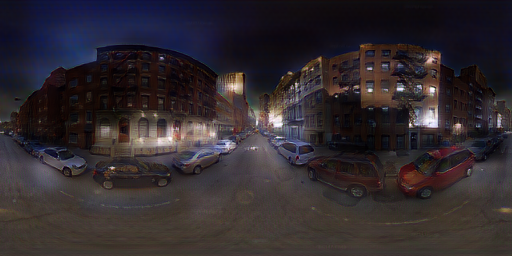}}
    \hfill
    \subfigure[]{\includegraphics[width= 0.33\linewidth]{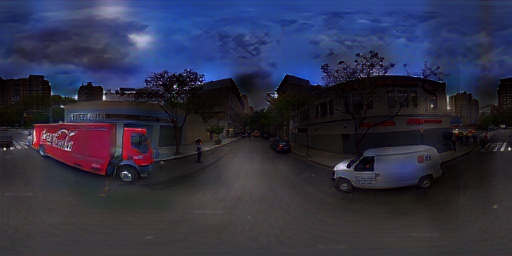}}
    \hfill
    \subfigure[]{\includegraphics[width= 0.33\linewidth]{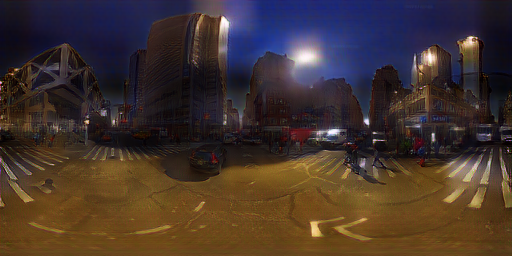}}
    \hfill \\\vspace{-10pt}
    
    {\small (b) Outputs from (a)} \\ \vspace{3pt}
  
    \subfigure[]{\includegraphics[width= 0.33\linewidth]{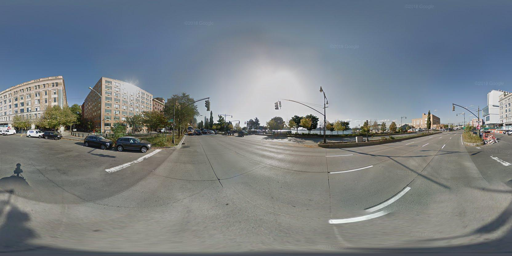}}
    \hfill
    \subfigure[]{\includegraphics[width= 0.33\linewidth]{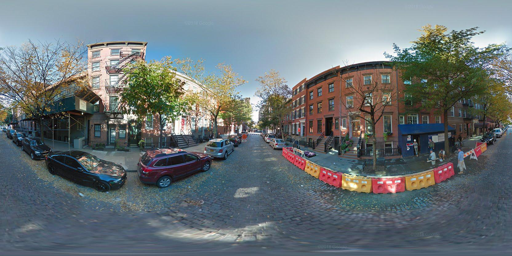}}
    \hfill
    \subfigure[]{\includegraphics[width= 0.33\linewidth]{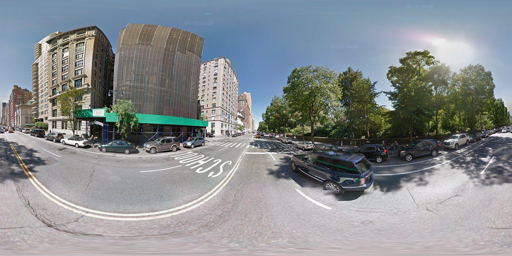}}
    \hfill \\\vspace{-10pt}
    
    {\small (c) Inputs} \\ \vspace{3pt}
    
    \subfigure[]{\includegraphics[width= 0.33\linewidth]{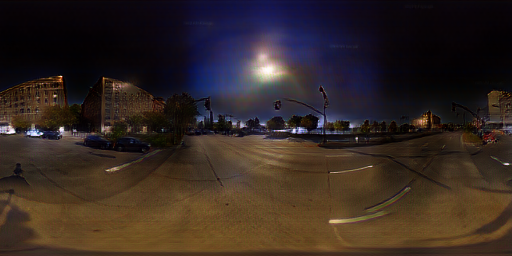}}
    \hfill
    \subfigure[]{\includegraphics[width= 0.33\linewidth]{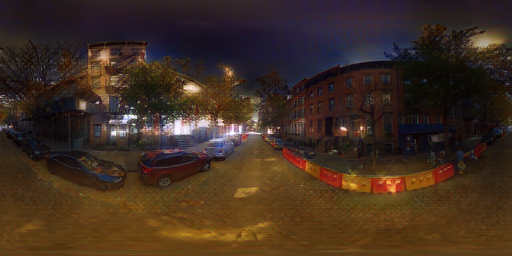}}
    \hfill
    \subfigure[]{\includegraphics[width= 0.33\linewidth]{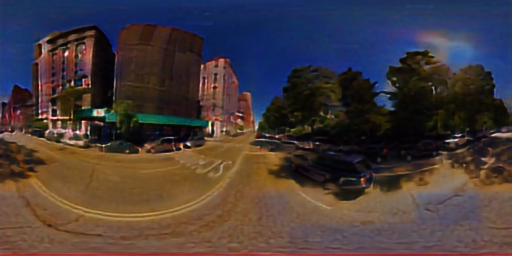}}
    \hfill \\\vspace{-10pt}
  
    {\small (d) Outputs from (c) } \\ \vspace{3pt}
    
    \caption{\textbf{Qualitative results} from \ours on StreetLearn dataset~\cite{mirowski2018learning} (day) to Dark Zurich dataset~\cite{shen2019towards} (twilight).}
    \label{fig:abl_qualitative_twilight}
 \end{center}
\end{figure*}

\begin{figure*}[h]
\begin{center}
\renewcommand{\thesubfigure}{}
    \subfigure[]{\includegraphics[width= 0.33\linewidth]{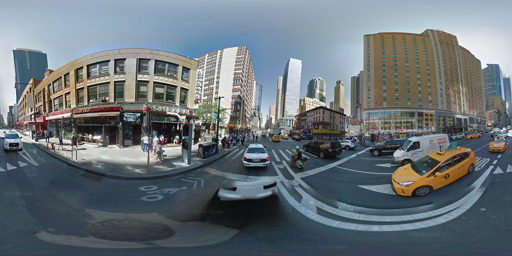}}
    \hfill
    \subfigure[]{\includegraphics[width= 0.33\linewidth]{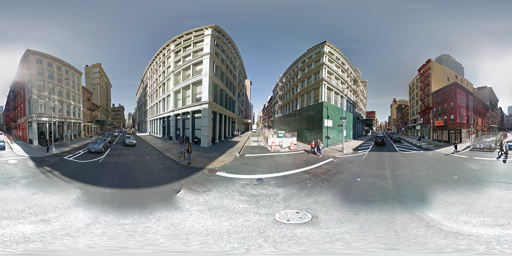}}
    \hfill
    \subfigure[]{\includegraphics[width= 0.33\linewidth]{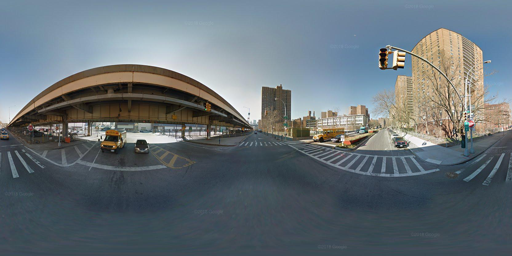}}
    \hfill \\\vspace{-10pt}
    
    {\small (a) Inputs} \\ \vspace{3pt}
  
    \subfigure[]{\includegraphics[width= 0.33\linewidth]{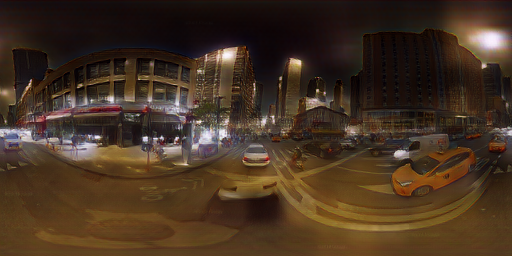}}
    \hfill
    \subfigure[]{\includegraphics[width= 0.33\linewidth]{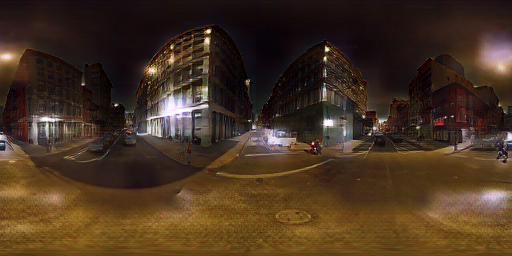}}
    \hfill
    \subfigure[]{\includegraphics[width= 0.33\linewidth]{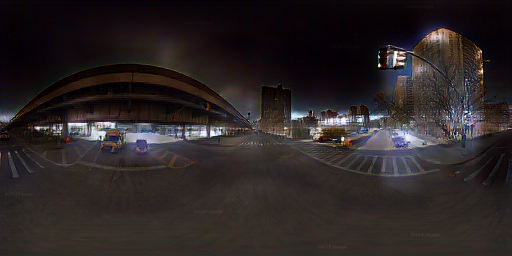}}
    \hfill \\\vspace{-10pt}
  
    {\small (b) Outputs from (a)} \\ \vspace{3pt}
    
    \subfigure[]{\includegraphics[width= 0.33\linewidth]{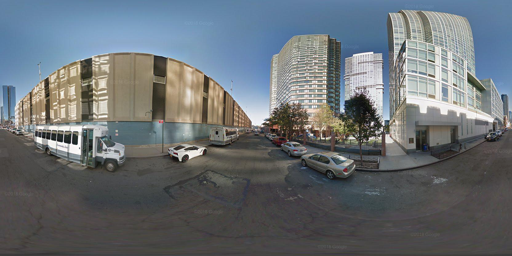}}  
    \hfill
    \subfigure[]{\includegraphics[width= 0.33\linewidth]{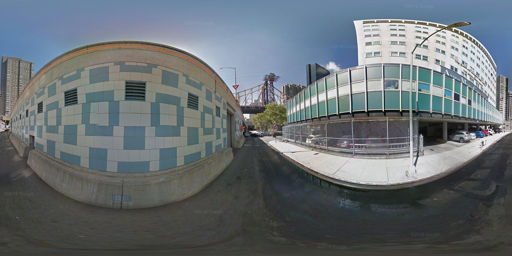}}
    \hfill  
    \subfigure[]{\includegraphics[width= 0.33\linewidth]{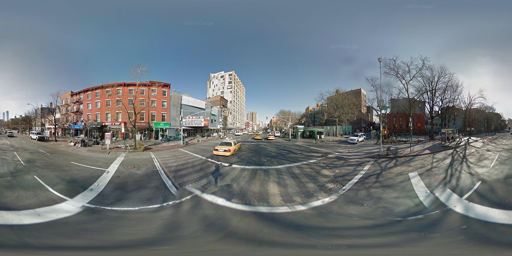}}
    \hfill \\\vspace{-10pt}
    
    {\small (c) Inputs} \\ \vspace{3pt}
    
    \subfigure[]{\includegraphics[width= 0.33\linewidth]{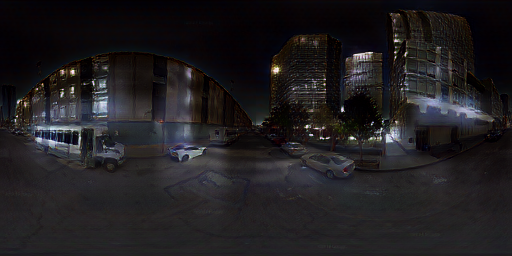}}
    \hfill  
    \subfigure[]{\includegraphics[width= 0.33\linewidth]{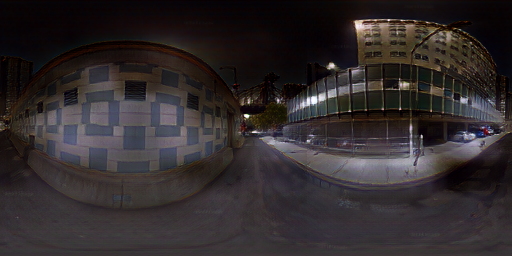}}  
    \hfill
    \subfigure[]{\includegraphics[width= 0.33\linewidth]{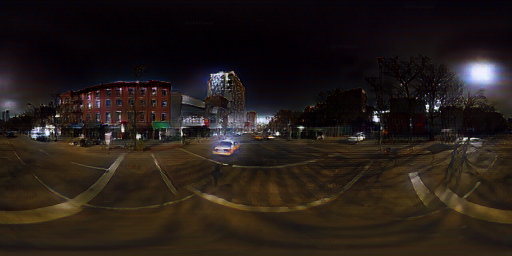}}
    \hfill\\\vspace{-10pt}
    
    {\small (d) Outputs from (c) } \\ \vspace{3pt}
    
    \caption{\textbf{Qualitative results} from \ours on StreetLearn dataset~\cite{mirowski2018learning} (day) to Dark Zurich dataset~\cite{shen2019towards} (night).}
  \label{fig:abl_qualitative_night}
\end{center}
\end{figure*}

\begin{figure*}
  \centering
  \renewcommand{\thesubfigure}{}
    \subfigure[]
{\includegraphics[width=0.249\linewidth]{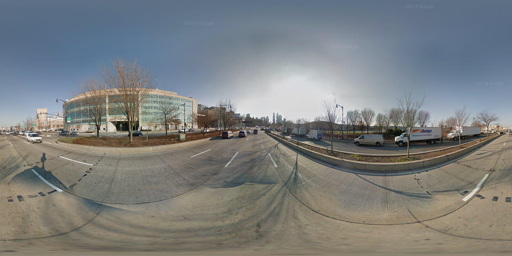}}\hfill
    \subfigure[]
{\includegraphics[width=0.249\linewidth]{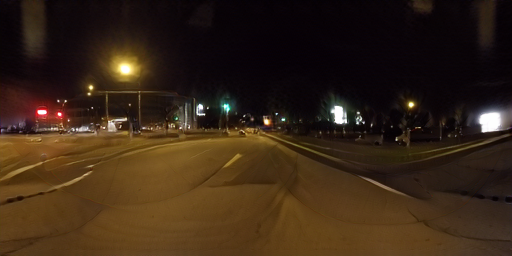}}\hfill
    \subfigure[]
{\includegraphics[width=0.249\linewidth]{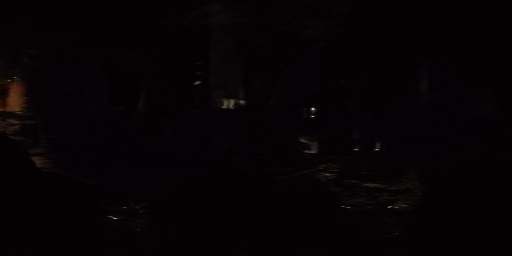}}\hfill
    \subfigure[]
{\includegraphics[width=0.249\linewidth]{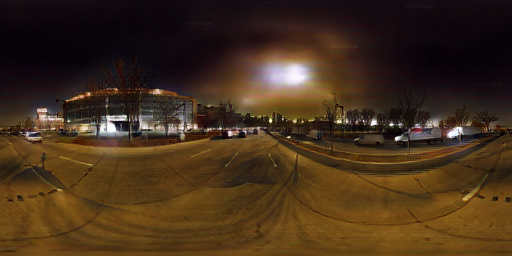}}\hfill\\\vspace{-20.5pt}

\subfigure[Inputs]
{\includegraphics[width=0.249\linewidth]{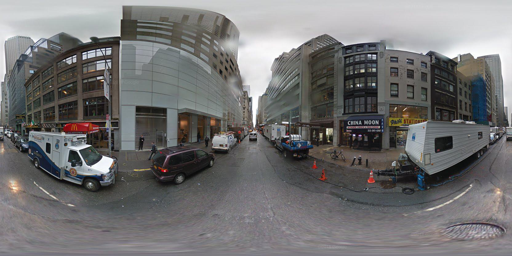}}\hfill
\subfigure[FSeSim~\cite{zheng2021spatially}]
{\includegraphics[width=0.249\linewidth]{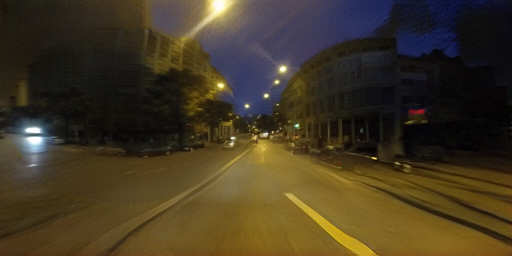}}\hfill
\subfigure[MGUIT~\cite{jeong2021memory}]
{\includegraphics[width=0.249\linewidth]{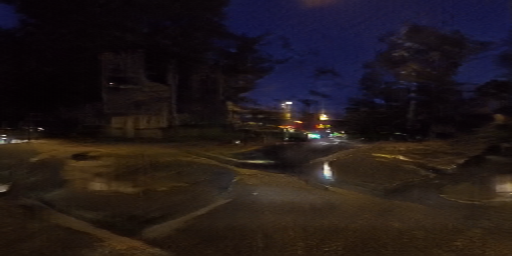}}\hfill
\subfigure[\ours (ours)]
{\includegraphics[width=0.249\linewidth]{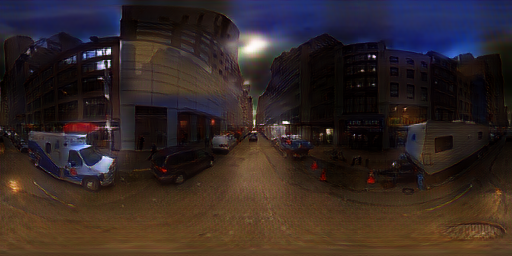}}\hfill\\
\caption{\textbf{Qualitative comparison} on StreetLearn dataset~\cite{mirowski2018learning} (day) to DZ dataset~\cite{sakaridis2019guided} (twilight, night): (top to bottom) day$\rightarrow$night, and day$\rightarrow$twilight results.}
  \label{fig:dz}
\end{figure*}

We visualize additional results of our method on another benchmark for the target domain, including day$\rightarrow$twilight in \figref{fig:abl_qualitative_twilight} and day$\rightarrow$night in \figref{fig:abl_qualitative_night} on StreetLearn and Dark Zurich~\cite{sakaridis2019guided}. We also provide visual comparisons with other methods~\cite{zheng2021spatially,jeong2021memory} in \figref{fig:dz}.
\vspace{15pt}

\clearpage
\newpage

\section{Details of user study}\label{supp:sec6} 
We conduct a user study to compare the subjective quality, shown in \figref{fig:userstudy} in the main paper. We randomly select 10 images for each task (sunny$\rightarrow$night, sunny$\rightarrow$rainy) for INIT dataset, compared with CUT~\cite{park2020contrastive}, MGUIT~\cite{jeong2021memory}, FSeSim~\cite{zheng2021spatially}, and InstaFormer~\cite{kim2022instaformer}.  
We request 60 participants to evaluate the quality of synthesized images, content relevance, and style relevance considering the context.
In particular, each instruction is as follows:

(1) Image quality. Given a row of 5 images, please select the index of the image that has the best image quality.

(2) Content relevance. Given a row of 5 images and a content image, please select the index of the image that has the most similar structure and content with A single content image.

(3) Style relevance considering the context. Given a row of 5 images, a content image, and a style image, please select the index of the image that has the most similar global style (weather, time, or color) to the style image.
Note that the images should have the same content as the content image.

\vspace{15pt}

\section{Limitations}\label{supp:sec7}
\begin{figure}[h]
  \centering
  \vspace{-5pt}
  \renewcommand{\thesubfigure}{}
    \subfigure[Input]{\includegraphics[width= 0.48\linewidth]
{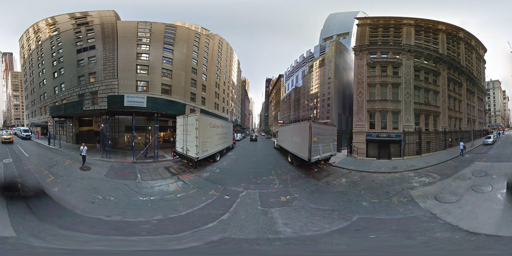}}
    \hfill
    \subfigure[Output (night)]{\includegraphics[width= 0.48\linewidth]{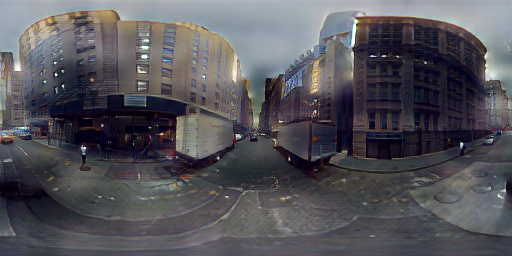}}
    \hfill \\
    \caption{\textbf{Failure case} of \ours on the StreetLearn dataset~\cite{mirowski2018learning} (day) to the INIT dataset~\cite{shen2019towards} (night).
    }
	\label{fig:supp_failure}
\end{figure}

Although our method shows outstanding performance on various benchmarks, our method inherits a problem in style relevance. 
Specifically, as observed in \figref{fig:supp_failure}, \ours is robust for preserving the panoramic structure, while sometimes struggling to represent the desired style for some samples. 
There is still room for performance improvement, so fostering research is needed.

\end{document}